\documentclass[acmtog]{acmart}
\acmSubmissionID{447}

\usepackage{booktabs} 
\usepackage{color}
\hypersetup{draft}
\setcopyright{none}
\renewcommand\footnotetextcopyrightpermission[1]{}

\usepackage[ruled]{algorithm2e} 

\SetAlFnt{\small}
\SetAlCapFnt{\small}
\SetAlCapNameFnt{\small}
\SetAlCapHSkip{0pt}

\acmJournal{TOG}

\begin{document}
\title{Attribute2Font: Creating Fonts You Want From Attributes}
\author{Yizhi Wang}
\authornote{Denotes equal contribution}
\affiliation{%
  \institution{Wangxuan Institute of Computer Technology, Peking University}
  \country{China}}
\email{wangyizhi@pku.edu.cn}

\author{Yue Gao}
\authornotemark[1]
\affiliation{%
  \institution{Wangxuan Institute of Computer Technology, Peking University}
  \country{China}}
\email{gerry@pku.edu.cn}
\author{Zhouhui Lian}
\authornote{Corresponding author}
\affiliation{%
  \institution{Wangxuan Institute of Computer Technology, Peking University}
  \country{China}}
\email{lianzhouhui@pku.edu.cn}

\begin{abstract}
Font design is now still considered as an exclusive privilege of professional designers, whose creativity is not possessed by existing software systems. Nevertheless, we also notice that most commercial font products are in fact manually designed by following specific requirements on some attributes of glyphs, such as italic, serif, cursive, width, angularity, etc. Inspired by this fact, we propose a novel model, Attribute2Font, to automatically create fonts by synthesizing visually pleasing glyph images according to user-specified attributes and their corresponding values. To the best of our knowledge, our model is the first one in the literature which is capable of generating glyph images in new font styles, instead of retrieving existing fonts, according to given values of specified font attributes. Specifically, Attribute2Font is trained to perform font style transfer between any two fonts conditioned on their attribute values. After training, our model can generate glyph images in accordance with an arbitrary set of font attribute values. Furthermore, a novel unit named Attribute Attention Module is designed to make those generated glyph images better embody the prominent font attributes. Considering that the annotations of font attribute values are extremely expensive to obtain, a semi-supervised learning scheme is also introduced to exploit a large number of unlabeled fonts. Experimental results demonstrate that our model achieves impressive performance on many tasks, such as creating glyph images in new font styles, editing existing fonts, interpolation among different fonts, etc.
\end{abstract}

%
%

%
%


\begin{teaserfigure}
  \includegraphics[width=\textwidth]{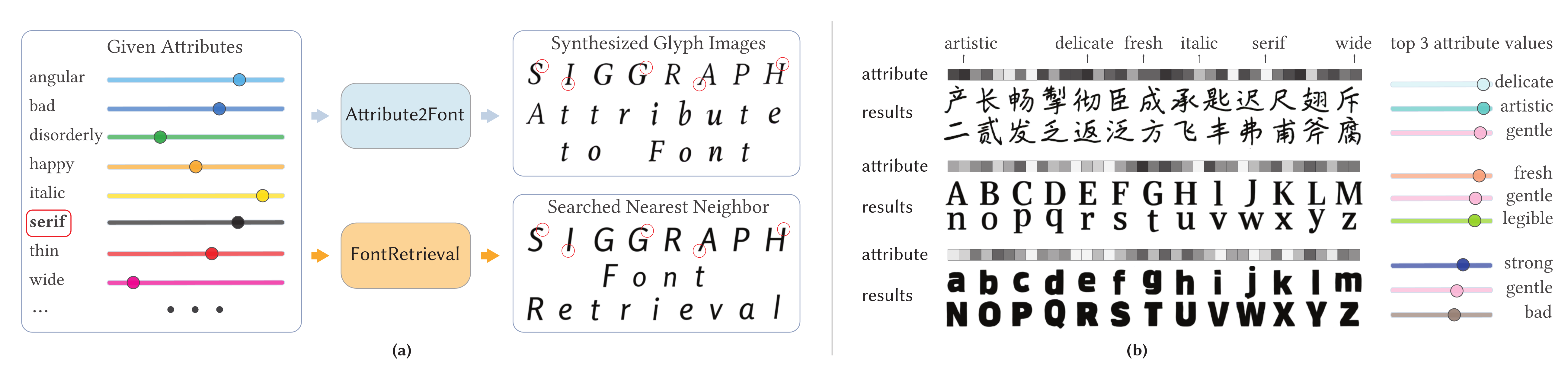}
  \caption{(a) An overview of our model, given an arbitrary set of predefined font attributes' values, glyph images in the corresponding style can be synthesized. By contrast, the font retrieval methods can only select existing fonts for users, which often cannot satisfy the specific requirements of some users. (b) Examples of synthesized English and Chinese glyph images in different attribute sets. We pre-define 37 different kinds of attributes whose values are shown in the grayscale grids (darker is higher). Please zoom in for better inspection.}
  \Description{.}
  \label{fig:teaser}
\end{teaserfigure}

\maketitle

\section{Introduction}

Traditional font design workflow sets a high barrier for common users, which requires creativity and expertise in this field.
Automatic font designing remains a challenging and ongoing problem in areas of Computer Graphics (CG), Computer Vision (CV) and Artificial Intelligence (AI).
In this paper, we aim to handle the task of generating glyph images according to the user-specified font attributes (such as italic, serif, cursive, angularity, etc.) and their values.
Our model significantly lowers the barrier and provides various and customizable fonts for common users.
We believe that our model can also inspire professional designers and assist them in creating new fonts.\par
To the best of our knowledge, our work is the first to automatically generate glyph images from the attributes of font styles.
Existing works on attribute-controllable image synthesis are unsuitable for this task due to the following three reasons:
(1) Existing works aim to generate images such as faces and fashion, whose appearances (such as color and texture) vary with the attributes but shapes generally remain unchanged.
By contrast, the shapes of glyphs vary dramatically with the font attributes.
(2) Existing works treat different attributes equally, whose effects are stacked into the generated images.
However, for each kind of font, some font attributes have much more impact on the glyph shapes than the other attributes.
Thereby, the correlation of different attributes should be explored and more attention should be paid to important attributes.
(3) The attribute values in these works are either binary (with or without) or discrete, which cannot be arbitrarily manipulated by users.\par
Up to now, large numbers of font generation methods have been reported intending to simplify the workflow of font designers.
There exist some works which attempt to ``create'' new fonts by manifold learning and interpolating between different fonts.
Nevertheless, these works fail to handle the situation when users have specific demands for the font style.
For example, one may seek for one kind of font which is attractive, sharp, with a little serif but not thin.
An alternative is employing font retrieval systems when they have such requirements.
However, these searching engines can only retrieve the most similar fonts from a font database that may not meet the users' needs (see Fig.~\ref{fig:teaser}).
In addition, the majority of them cannot accept attributes with values as keywords.\par
To address the above-mentioned problems, this paper proposes a novel model which is capable of generating glyph images according to user-specified font attributes, named \textbf{Attribute2Font}~\footnote{Source code is available at \url{https://hologerry.github.io/Attr2Font/}}.
In the first place, we assume that each font corresponds to a set of font attribute values.
On the basis of this assumption, we train a deep generative network to transfer glyph images in a font style to another according to their font attribute values.
In the inference stage, we choose an appropriate font as source and transform it into any fonts the users want from given attribute values.
Technically, we propose a semi-supervised learning scheme and the Attribute Attention Module to boost our model's performance.
The semi-supervised learning scheme is introduced to deal with the shortage of annotated training data.
The Attribute Attention Module assists our model to concentrate on the most important attributes and better portray the prominent characteristics of glyphs in the generation stage.
Experimental results on publicly available datasets demonstrate the effectiveness of our model in many applications, such as creating glyph images in new font styles, editing existing fonts, interpolation among different fonts, etc.
Last but not least, our experiments verify that our model can also be applied to deal with fonts in any other writing systems (e.g., Chinese) which might contain hundreds of thousands of different characters.\par
To sum up, major contributions of this paper are as follows:
\begin{itemize}
\item For the first time, we establish a mapping relation from descriptive font attributes to the glyph image space. Users are allowed to arbitrarily set values of the predefined font attributes and create any fonts they want.
\item The devised semi-supervised learning scheme and attribute attention module significantly promote the quality of generated glyph images.
Besides, they are not only limited to our task but also applicable to other image synthesis tasks.
\item Our model is capable of synthesizing highly varied and visually pleasing glyph images. Therefore, our model has a high practical value in font generation for both ordinary users and font designers.
\end{itemize}
\par
\section{Related Work}
\subsection{Attribute-controllable Image Synthesis}
Image synthesizing methods essentially fall into two categories: parametric and non-parametric.
The non-parametric methods generate target images by copying patches from training images.
In recent years, the parametric models based on Generative Adversarial Networks (GANs)~\cite{goodfellow2014generative} have been popular and achieved impressive results.
Image-to-image translation is a specific problem in the area of image synthesis.
Most recent approaches utilize CNNs (Convolutional Neural Networks) to learn a parametric translation function by training a dataset of input-output examples.
Inspired by the success of GANs in generative tasks, the ``pix2pix'' framework~\cite{isola2017image} uses a conditional generative adversarial network to learn a mapping from input to output images. \par
Many recent works utilize GAN-based models to synthesize attribute-controllable images, such as face images with controllable hair colors, ages, genders, etc.
Xu et al.~\shortcite{xu2018attngan} directly mapped the descriptive texts into images by combining the techniques of natural language processing and computer vision.
Choi et al.~\shortcite{choi2018stargan}, He et al.~\shortcite{he2019attgan}, Liu et al.~\shortcite{liu2019stgan} and Wu et al.~\shortcite{wu2019relgan} achieved this goal by implementing image-to-image translation conditioned on image attributes.
Compared to the direct mapping from attributes to images, image-to-image translation usually generates more realistic images on the basis of the strong prior of source images.
However, there is no existing work attempting to synthesize glyph images based on font style attributes.
Besides, the attribute values in existing works are discrete rather than continuous, which constrains the variety and diversity of synthesized images.
\begin{figure*}[t!]
  \centering
  \includegraphics[width=\textwidth]{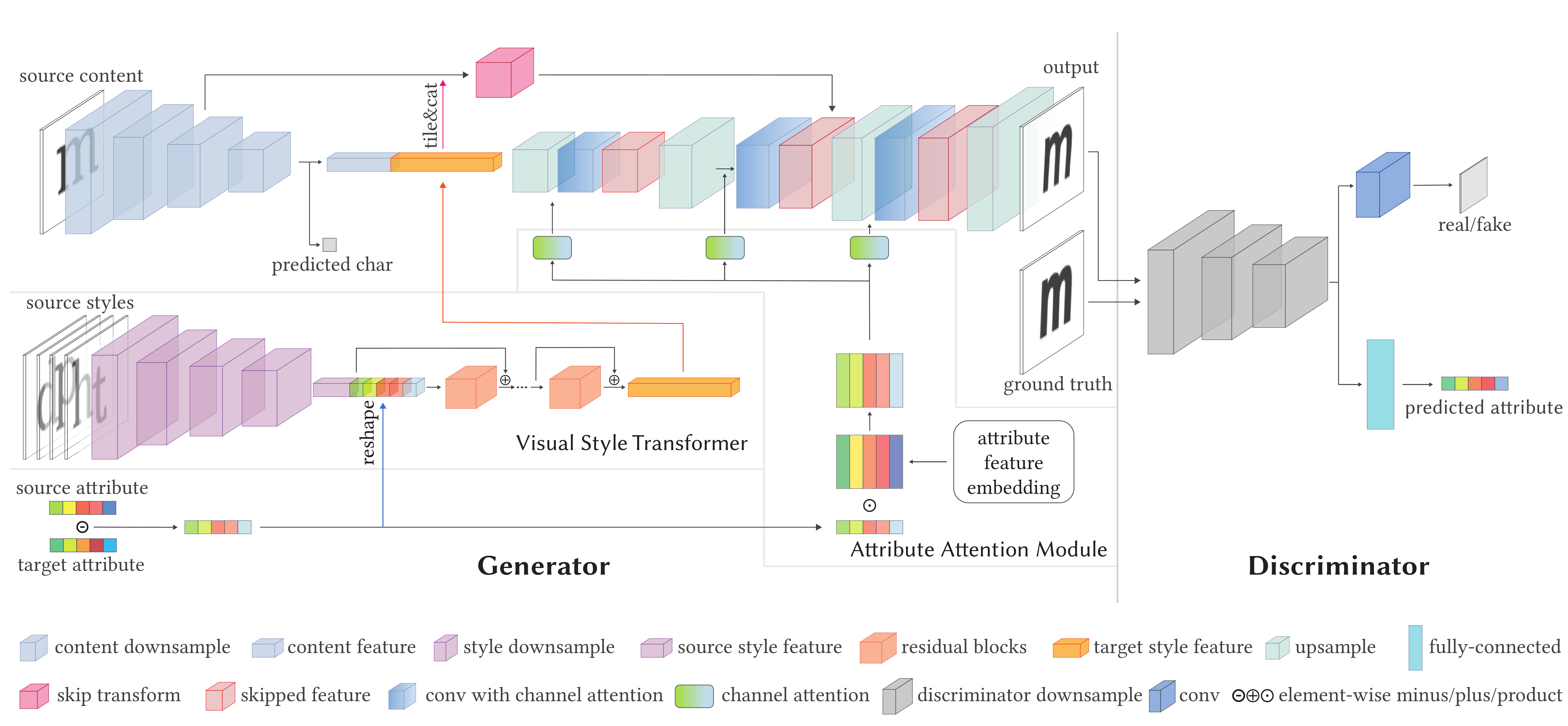}
  \caption{The pipeline of our proposed method.}
  \label{fig:Pipeline}
\end{figure*}
\subsection{Font Generation}
Existing works of font generation have two main practical applications.
The first one is simplifying the workflow of font designers.
They typically take a few (or part) glyphs of a font as reference and then generate the rest glyphs in this font.
They fully exploit the shared stylistic elements within the same font and the similar structure that character glyphs present across different fonts.
The second one is ``creating" new fonts by integrating and mixing up existing font styles via interpolation or/and manifold learning methods.
This benefits the designers and users who desire to freely customize their favourite font styles.
In terms of the representation of generated glyphs, these works can be classified into two categories: vector font generation and glyph image synthesis.
\subsubsection{Vector Font Generation}
Generating vector fonts is a typical problem in the area of CG.
Campbell and Kautz~\shortcite{campbell2014learning} built a font manifold and generated new fonts by interpolation in a high dimensional space.
Lian et al.~\shortcite{lian2018easyfont} proposed a system to automatically generate large-scale Chinese handwriting fonts by learning styles of stroke shape and layout separately.
Balashova et al.~\shortcite{balashova2019learning} proposed to learn a stroke-based font representation from a collection of existing typefaces.
Lopes et al.~\shortcite{lopes2019learned} attempted to model the drawing process of fonts by building sequential generative models of vector graphics.
However, vector font generation suffers from the contradiction between various and flexible font styles and structured representations of visual concepts.
For example, the model proposed by Campbell and Kautz~\shortcite{campbell2014learning} is limited by the need of having all glyphs of a certain class be composed of equivalent shapes;
When trained on a large-scale font dataset, the model presented by Lopes et al.~\shortcite{lopes2019learned} generates vector glyphs whose quality is far from satisfactory.
\subsubsection{Glyph Image Synthesis}
Glyph image synthesis loses the benefit of providing a scale-invariant representation but is able to generate glyphs in complex shapes and novel styles.
For example, Lyu et al.~\shortcite{lyu2017auto} devised a model to synthesize Chinese calligraphy.
Azadi et al.~\shortcite{azadi2018multi} proposed an end-to-end stacked conditional GAN model to generate a set of multi-content glyph images following a consistent style from very few examples.
Guo et al.~\shortcite{guo2018creating} proposed a method to synthesize Chinese glyph images in new font styles by building font manifolds learned from existing font libraries.
Jiang et al.~\shortcite{jiang2017dcfont,jiang2019scfont} and Gao et al.~\shortcite{gao2019artistic} viewed font style transfer as an
image-to-image translation problem and mapped glyph images from one font style to another style.
Given a few glyph image pairs of the source and target fonts, the deep generative networks they utilized learn a mapping function and then transfer the other glyph images in the source font style to the target font style.
However, the mapping function is abstract and users are not allowed to get their desired fonts by specifying their expected attributes.
To remedy this issue, we design an attribute-controllable font style transfer model where users can customize the attribute categories and arbitrarily assign the attribute values.

\subsection{Font Selection and Retrieval}
Font selection and retrieval is about how to select fonts from a font database according to the conditions provided by users.
Wang et al.~\shortcite{wang2015deepfont} designed the DeepFont system which is queried by a real-world text image and retrieves the most similar fonts from a font library.
O’Donovan et al.~\shortcite{o2014exploratory} first proposed to use high-level descriptive attributes for font selection, such as ``dramatic" or ``legible."
Choi et al.~\shortcite{choi2019assist} and Chen et al.~\shortcite{chen2019large} addressed the task of large-scale tag-based font retrieval, employing deep neural networks to build the bridge between descriptive tags and glyph images in various fonts.
However, these works are merely aimed at searching existing fonts for users, which limits the creativity and flexibility of their models and may not satisfy the needs of users.
In this paper, we propose to directly generate glyph images from given descriptive attributes.
Our model receives continuous attribute values as input instead of binary tags in~\cite{chen2019large}.
Users are allowed to control the value of each attribute and get the glyph images they want.

\section{Method Description}
\subsection{Overview}
\label{sec:method_des_ov}
We first briefly describe the pipeline of Attribute2Font shown in Fig.~\ref{fig:Pipeline}.
Generally, our model aims to transfer the glyph images in one font style to another according to their font attributes.
Let $x(a,k)$ and $x(b,k)$ be the glyph image pair with the same $k$-th character category but in the $a$-th and $b$-th fonts, respectively,
where $1 \leq k \leq N_{c}$ and $1 \leq a,b \leq N_{f}$, $N_{c}$ and $N_{f}$ are the total numbers of character categories and fonts in a database, respectively.
$x(a,k)$ and $x(b,k)$ denote the source and target in the transfer, respectively, and are marked as ``source content'' and ``ground truth'' in Fig.~\ref{fig:Pipeline}.
The glyphs in the same font share the same font attribute.
Let $\alpha(a)$ and $\alpha(b)$ be the attribute values of $x(a,k)$ and $x(b,k)$, respectively.
Let $N_{\alpha}$ be the category number of predefined attributes and thus $\alpha(a),\alpha(b) \in \mathbb{R}^{N_{\alpha}}$.
Note that the categories of font attributes can be freely customized but they remain fixed when the model is deployed.
There is no requirement of these attributes being independent of each other.
In principle the attribute values can be arbitrarily assigned but need to be normalized into the range of $[0,1]$ before sent into the model (`0' denotes minimum and `1' denotes maximum).
\par
The Visual Style Transformer (VST) aims to estimate the style feature of the $b$-th font on the basis of glyph images of the $a$-th font and the attribute difference between these two fonts.
A few glyph images in the font $a$ but different character categories are taken as input, denoted as $\{x(a,k_{1}), ...,x(a,k_{m})\}$ and presented as ``source styles'' in Fig.~\ref{fig:Pipeline}, where
$k_{1}, ..., k_{m}$ are randomly sampled from $\{1, ..., N_{c}\}$ and $m$ is a hyperparameter denoting the number of these glyph images.
The estimated style feature of font $b$ is formulated as:
\begin{equation}
\label{equ:GlyGen}
 \hat{s}(b) = F_{T}(x(a,k_{1}), ...,x(a,k_{m}),\alpha(b)-\alpha(a)),
\end{equation}
where $F_{T}$ denotes the function of VST.\par
The Attribute Attention Module (AAM) is devised to further refine the attributes so that they can serve better in the stage of glyph image generation:
\begin{equation}
\label{equ:GlyGen}
 \alpha^{*}(a,b) = F_{A}(\alpha(b) - \alpha(a)),
\end{equation}
where $\alpha^{*}$ denotes the refined attribute difference by applying the attention mechanism and $F_{A}$ denotes the function of AAM. \par

The Generator takes the source image $x(a,k)$, the estimated style feature $\hat{s}(b)$, and the refined attribute difference $\alpha^{*}(a,b)$ as input to reconstruct the glyph image ${x}(b,k)$:
\begin{equation}
\label{equ:GlyGen}
 \hat{x}(b,k) = G(x(a,k), \hat{s}(b), \alpha^{*}(a,b)),
\end{equation}
where $G$ denotes the function of glyph generator. \par
Following the adversarial training scheme of GANs, we employ a glyph discriminator to discriminate between the generated image and the ground-truth image, i.e., $\hat{x}(b,k)$ and $x(b,k)$.
The glyph discriminator takes $x(b,k)$ or $\hat{x}(b,k)$ as input and predicts the probability of the input image being real (denoted as $p(y_{d} = 1)$) and its corresponding attributes $\hat{\alpha}$:
\begin{equation}
\label{equ:GlyGen}
 p(y_{d} = 1 | x(b,k)), \hat{\alpha}(x(b,k)) = D( x(b,k)),
\end{equation}
\begin{equation}
\label{equ:GlyGen}
p(y_{d} = 1 | \hat{x}(b,k)), \hat{\alpha}(\hat{x}(b,k)) = D(\hat{x}(b,k)).
\end{equation}
Through the adversarial game between the generator and the discriminator, the quality of generated glyphs can be continuously improved.
More details of above-mentioned modules utilized in our method are presented in the following sections.
\subsection{Visual Style Transformer}
We employ a CNN style encoder to transform the selected glyph images of the $a$-th font into its style feature which is denoted as $s(a)$:
\begin{equation}
s(a) = F_{S}([x(a,k_{1}); ...;x(a,k_{m})]),
\end{equation}
where the square bracket denotes concatenation, in other words, all images are concatenated along the depth channel and then fed into the style encoder (see Fig.~\ref{fig:Pipeline}).
$F_{S}$ denotes the function of style encoder.
Empirically, a single glyph is unable to sufficiently embody its belonging font style.
Therefore, we set $m \textgreater 1$ for more accurate estimation of $s(a)$.
In Section~\ref{sec:experiments}, we will show experimental results that demonstrate how the setting of $m$ affects our model's performance.
Next, the encoded feature $s(a)$ is concatenated with the attribute difference between the attribute values of $x(a,k)$ and $x(b,k)$, i.e., $\alpha(b) - \alpha(a)$.
Afterwards we send them into several residual blocks~\cite{he2016deep} and finally have the estimated target style features:
\begin{equation}
\hat{s}(b) = F_{R}([s(a);\alpha(b) - \alpha(a)]),
\end{equation}
where $F_{R}$ denotes the function of residual blocks.
The number of residual blocks is denoted as $N_{rb}$ and its effect on our method will also be investigated in Section~\ref{sec:experiments}.
\subsection{Attribute Attention Module}
\begin{figure}[t!]
  \centering
  \includegraphics[width=8cm]{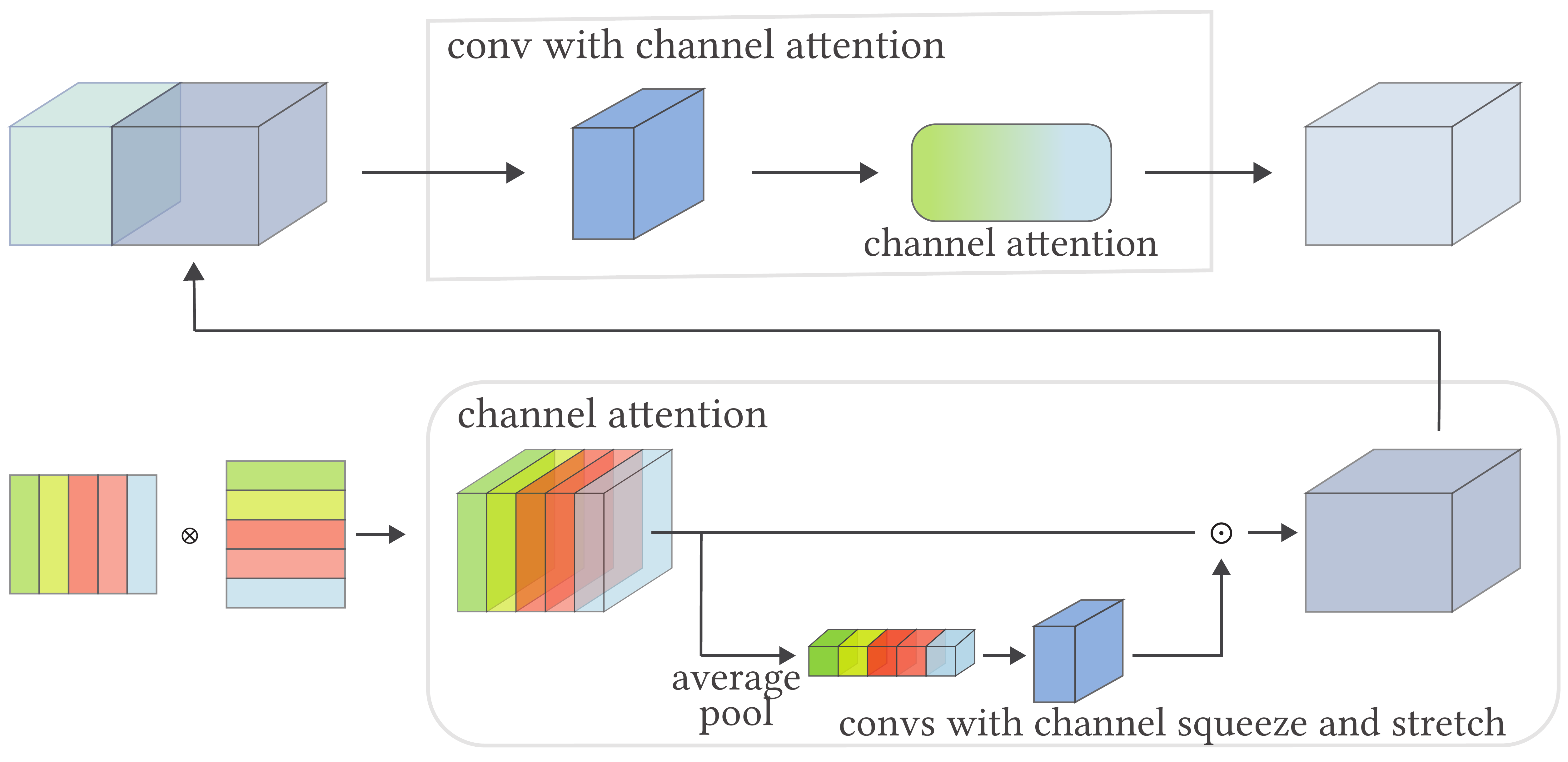}
  \caption{The architecture of our Attribute Attention Module and how it is incorporated into the feature maps in the decoder.}
  \label{fig:channel_attention}
\end{figure}

In this section, we give a detailed description of the Attribute Attention Module (AAM) which is illustrated in Fig.~\ref{fig:channel_attention}.
We introduce attribute embeddings $e \in \mathbb{R}^{N_{\alpha} \times N_{e}}$ as more concrete representations of the features of each attribute, where $N_{e}$ is interpreted as the dimension of attribute embeddings.
The attribute values $\alpha$ could be viewed as the weight coefficients of attribute embeddings.
The attribute feature difference between the $b$-th and $a$-th fonts is represented as:
\begin{equation}
\beta(a,b)=  (\alpha^{t}(b) - \alpha^{t}(a)) \odot e,
\end{equation}
where $\odot$ denotes the element-wise multiplication; $\alpha^{t}(b)$, $\alpha^{t}(a) \in \mathbb{R}^{N_{\alpha} \times N_{e}}$ are tiled from $\alpha(b)$ and $\alpha(a)$, respectively; $\beta(a,b) \in \mathbb{R}^{N_{\alpha} \times N_{e}}$.
We map the feature vector of each attribute in $\beta(a,b)$ into a two-dimensional feature map by:
\begin{equation}
\gamma(a,b) = \beta_{0}(a,b) \otimes \beta'_{0}(a,b),
\end{equation}
where $\beta_{0}(a,b) \in \mathbb{R}^{N_{\alpha} \times N_{e} \times 1}$ is unsqueezed from $\beta(a,b) \in \mathbb{R}^{N_{\alpha} \times N_{e}} $; $\beta'_{0}(a,b)\in \mathbb{R}^{N_{\alpha} \times 1 \times N_{e}}$ is the transpose of $\beta_{0}(a,b)$; $\otimes$ denotes the matrix multiplication over the last two dimensions and thus we have $\gamma(a,b) \in \mathbb{R}^{N_{\alpha} \times N_{e} \times N_{e}}$.
The feature map for each attribute in $\gamma(a,b)$ is a symmetric matrix and is in favour of convolution operations in the following stage.
Afterwards, we perform channel attention on $\gamma(a,b)$:
\begin{equation}
\alpha^{*}(a,b) = F_{CA}(\gamma(a,b)),
\end{equation}
where $F_{CA}$ denotes the channel attention block.
The channel attention operation was first proposed in~\cite{woo2018cbam} and~\cite{zhang2018image}, which produces a channel attention map by exploiting the inter-channel relationship of features.
Specifically, the feature maps in $\gamma(a,b)$ are first aggregated by an average pooling layer and then sent into two convolution layers with channel squeeze and stretch to output the channel attention map $M(a,b) \in \mathbb{R}^{N_{\alpha} \times 1 \times 1}$.
$\alpha^{*}(a,b)$ is computed by the channel-wise multiplication between $M(a,b)$ and $\gamma(a,b)$.
We utilize the channel attention to explore the correlation of different attributes and help our model concentrate on more important attributes.
The attribute attention is performed on the different stages of glyph generation, and the output will be re-scaled into the desired scale of each stage (details will be revealed in the next section).
\subsection{Glyph Image Generation}
A hierarchical encoder-decoder framework is deployed to generate the target glyph image (i.e., $x(b,k)$) on the basis of $\hat{s}(b)$, $\alpha^{*}(a,b)$ and $x(a,k)$.
The encoder, named as the content encoder, is a CNN which progressively maps $x(a,k)$ into its content feature $c(a,k)$.
Multi-scale features from the content encoder are utilized to more accurately reconstruct the glyph's shape, which is shown in the skip-connection in Fig.~\ref{fig:Pipeline}.
Note that only one example of skip-connection is demonstrated for brevity.
We denote them as $c_{1}(a,k)$, $c_{2}(a,k)$, ..., $c_{L}(a,k)$ and $L$ is the number of feature scales.
A softmax classifier is attached to $c_{L}(a,k)$ to predict the character category of $x(a,k)$.
The probability of $x(a,k)$ belonging to the ${k}$-th category is denoted as $p(y_{c} =k|x(a,k))$.
We also have $L$ up-sampling layers in the decoding step.
The output of the $i$-th up-sampling layer in the decoder is formulated as:
\begin{equation}
g_{i}=\begin{cases}F_{u}([F_{CA}([g_{i-1};\alpha^{*}_{i-1}]);h_{i-1}]), &2 \leq i \leq L \cr F_{u}(h_{i-1}), &i = 1\end{cases},
\end{equation}
where $F_{u}$ is the function of up-sampling, including deconvolution, instance normalization~\cite{ulyanov2016instance} and activation operations;
$\alpha^{*}_{i-1}$ denotes the re-scaled output of AAM in the $(i-1)$-th decoding stage;
$h_{i}$ is the fusion of ${c}_{L-i}(a,k)$ and $\hat{s}(b)$.
Specifically, $\hat{s}(b)$ is first tiled according to the shape of ${c}_{L-i}(a,k)$ and then concatenated with ${c}_{L-i}(a,k)$:
\begin{equation}
h_{i}=\begin{cases}F_{c}([\hat{s}_{i}^{t}(b);{c}_{L-i}(a,k)]), &1 \leq i \leq L-1 \cr [\hat{s}(b);{c}_{L-i}(a,k)], &i = 0\end{cases},
\end{equation}
where $F_{c}$ denotes the function of convolution and $\hat{s}_{i}^{t}(b)$ is tiled from $\hat{s}(b)$.
We have $\hat{x}(b,k) = g_{L}$ and the last up-sampling layer uses $tanh$ as the activation function.
\subsection{Loss Functions}
We define five losses in the step of generation.
The pixel-level loss $l_{pixel}$ measures the dissimilarity between $\hat{x}(b,k)$ and ${x}(b,k)$ in the level of pixel values using the L1 Norm:
\begin{equation}
l_{pixel} =  \Vert \hat{x}(b,k) - {x}(b,k)\Vert.
\end{equation}
The content loss $l_{char}$ is the cross-entropy loss for character recognition:
\begin{equation}
l_{char} = -\log p( y_{c} = k | x(a,k)).
\end{equation}
$l_{attr}$ is the prediction loss of the discriminator predicting the attribute values of generated images:
\begin{equation}
l_{attr} = smooth_{L_{1}}(\hat{\alpha}(\hat{x}(b,k))- \alpha(b)),
\end{equation}
and the $smooth_{L_{1}}$ function is defined as:
\begin{equation}
smooth_{L_{1}}(x)=\begin{cases}0.5x^{2}, &\left |x \right |\leq 1 \cr \left |x \right|-0.5, &otherwise\end{cases}.
\end{equation}
The contextual loss was recently proposed in~\cite{mechrez2018contextual}.
It is a new and effective way to measure the similarity between two images, requiring no spatial alignment.
As the spatial alignment is required for the L1 loss, if the synthesized image is not exactly spatially aligned to the ground truth image (e.g., a small displacement or rotation), the L1 loss will be high but the synthesis result is often visually acceptable.
The contextual loss leads the model to pay more attention to style features at a high
level, not just differences in pixel values.
Therefore, we regard the contextual loss as a complementary to the L1 loss:\par
\begin{equation}
l_{CX} = CX(\hat{x}(b,k), {x}(b,k)).
\end{equation}
The vanilla generation loss $l_{G}$ is defined as:
\begin{equation}
l_{G} = -\log p(y_{d} = 1 | \hat{x}(b,k)).
\end{equation}
The total loss of the generation step is formulated as:
\begin{equation}
L_G = \lambda_{1} l_{G} + \lambda_{2} l_{pixel} + \lambda_{3} l_{char} + \lambda_{4} l_{CX} + \lambda_{5} l_{attr},
\end{equation}
where $\lambda_{1},\lambda_{2},...,\lambda_{5}$ are hyperparameters denoting the weights of loss functions.
\par
In the step of discrimination, we define two losses.
The first one $l'_{attr}$ is the prediction loss of the discriminator that predicts the attribute values of ground-truth images:
\begin{equation}
l'_{attr} =  smooth_{L_{1}}(\hat{\alpha}(x(b,k))- \alpha(b)).
\end{equation}
The other one, the vanilla discrimination loss, $l_{D}$ is formulated as:
\begin{equation}
l_{D} = -\log p(y_{d} = 1 | x(b,k)) -\log p(y_{d} = 0 | \hat{x}(b,k)).
\end{equation}
The total loss of the discrimination step is formulated as:
\begin{equation}
L_D = l_{D} + l'_{attr}.
\end{equation}
We optimize these two objective functions ($L_{G}$ and $L_{D}$) alternately.
\subsection{Semi-supervised Learning}
~\cite{o2014exploratory} released a font dataset consisting of 1,116 fonts where the attribute values of 148 fonts are annotated.
Such a small quantity of training data is far from enough to train a deep neural network with satisfactory performance.
Nevertheless, they paid a huge effort to obtain these annotations by sending questionnaires to the Mechanical Turk and sum them up by machine learning models.
To remedy this issue, we propose a semi-supervised learning scheme to exhaustively exploit unlabeled fonts.
Our main idea is to incorporate the annotating of unlabeled fonts’ attribute values into the font style transfer process.
The glyph images in a training image pair could be either selected from the labeled fonts or the unlabeled fonts.
We first assign pseudo attribute values for the unlabeled fonts and they will be fine-tuned by the gradient descent algorithm.
In this manner, our model learns to annotate the attribute values of unlabeled fonts by referring to the distribution of human-annotated attributes.
Details of our semi-supervised learning scheme are described as follows.
\par
\begin{figure}[t!]
  \centering
  \includegraphics[width=8cm]{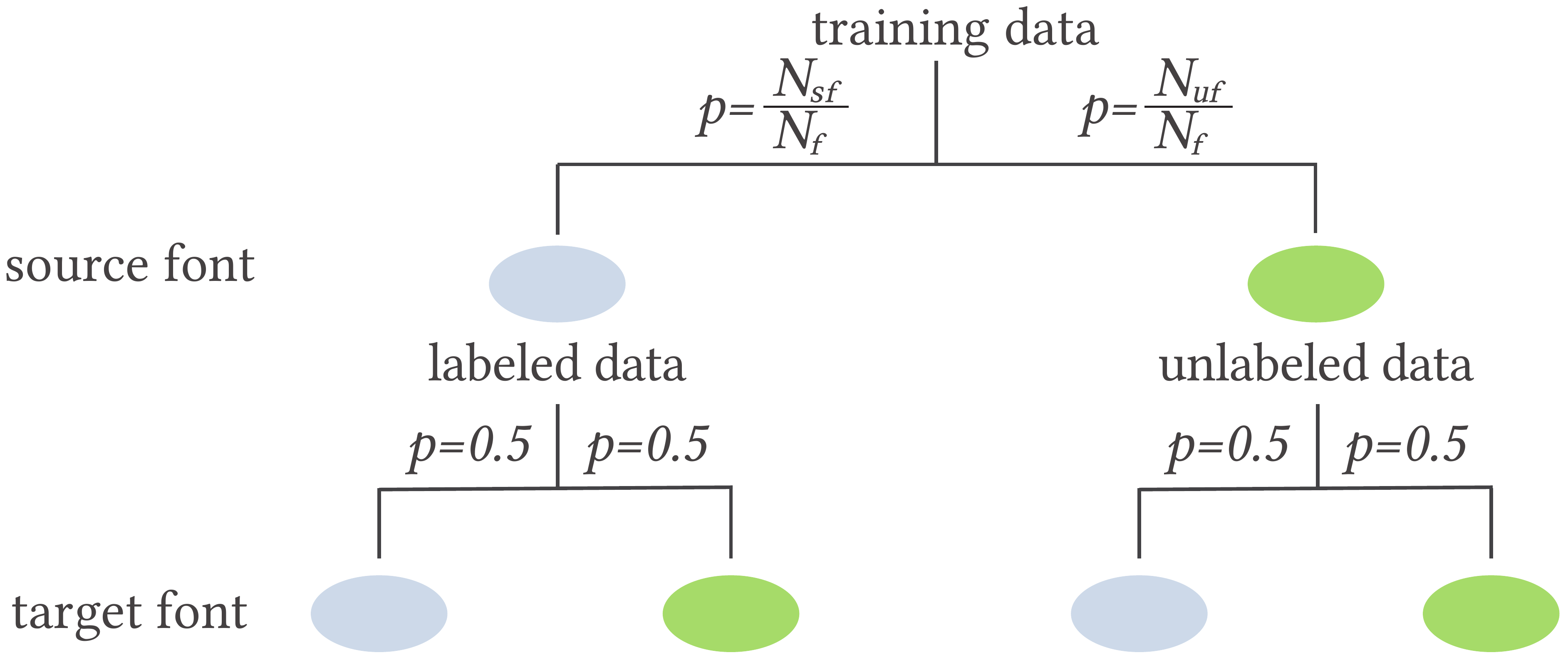}
  \caption{How we select from the labeled and unlabeled fonts to make image pairs for training.}
  \label{fig:training_strategy}
\end{figure}

Assume that we have $N_{sf}$ fonts whose attribute values are labeled and $N_{uf}$ fonts whose attribute values are unlabeled ($N_{uf} + N_{sf} =  N_{f}$ and $N_{uf} \gg N_{sf}$, typically).
Let $\Phi{s}$ consist of the indexes of labeled fonts while $\Phi{u}$ consist of the indexes of unlabeled fonts.
The font of the source image is randomly selected, i.e., the probability of the source image being in a labeled font $p(a \in \Phi{s}) = \frac{ N_{sf}}{N_{sf} + N_{uf}}$  and being in a unlabeled font $p(a \in \Phi{u}) = \frac{ N_{uf}}{N_{sf} + N_{uf}}$.
For the target image, selecting from the labeled fonts and unlabeled fonts are equally likely, i.e., $p(b \in \Phi{s}) = 0.5$ and $p(b \in \Phi{u}) = 0.5$.
We increase the proportion of labeled fonts in target images compared to source images because the strongest supervision (such as $l_{pixel}$, $l_{CX}$) comes from the target image.
The details of this combination strategy are illustrated in Fig~\ref{fig:training_strategy}.
We first randomly initialize the attribute values of unlabeled fonts according to the standard Gaussian distribution $\mathcal N(0,1)$ attached by the sigmoid function:
\begin{equation}
\alpha(i) = sigmoid(z), i \in \Phi{u},
\end{equation}
where $z \sim \mathcal N(0,1)$; $sigmoid$ is the sigmoid function which maps $z$ into the range of $(0,1)$.
In the training phase, the attribute values of labeled fonts are fixed but the ones of unlabeled fonts are fine-tuned via the gradient descent algorithm.
Namely, if $a$ or $b \in \Phi{s}$, $\alpha(a)$ or $\alpha(b)$ remains fixed; If $a$ or $b$ $\in \Phi{u}$, $\alpha(a)$ or $\alpha(b)$ is fine-tuned by using the gradient descent algorithm.
The scheme is proved to be very effective through our experiments, which will be further discussed in Section~\ref{sec:experiments}.
\par
\subsection{The Choice of Source Font in Inference Stage}
In the training stage, our model performs font style transfer between any two fonts in the training dataset.
But in the inference stage, our model is only given a set of attribute values.
This raises an issue of how to select an appropriate source font so that it can be easily transferred into the desired font.
Since we estimate the style feature $\hat{s}(b)$ from the glyph images of the source font,
$\hat{s}(b)$ is fused with the source font's style more or less.
Intuitively, selecting a source font which is similar with the desired font results in good performance; selecting a source font which differs a lot against the desired font results in bad performance.
The similarity is measured by comparing the attribute values of the user's input and the candidate source font.
However, our experiments described in Section~\ref{sec:experiments} demonstrate that the generated glyph images are nearly the same for most source fonts.
Eventually, we draw the conclusion that there is no strict restriction on the source font in our model.

\section{Experiments}
\label{sec:experiments}
\subsection{Dataset}
We adopt the font dataset released by~\cite{o2014exploratory} which consists of 148 attribute-labeled fonts and 968 attribute-unlabeled fonts.
For convenience, we refer this dataset as ``AttrFont-ENG''.
There are 37 kinds of font attributes ($N_{\alpha} = 37$), containing both concrete attributes such as “thin” and “angular,” and more nebulous concepts like “friendly” and “sloppy”.
The annotation task was done by asking Mechanical Turk workers to compare different fonts according to their attributes and estimating relative scalar values for each font.
Among these labeled fonts, we set the first 120 fonts as our supervised training data, the last 28 fonts as our validation data.
Namely, $N_{sf} = 120$ and  $N_{uf} = 968$.
In this dataset, each font is represented by 52 glyph images (a-b and A-B), i.e., $N_{c} = 52$.

\subsection{Implementation Details}
The proposed model is implemented in PyTorch and trained on a NVIDIA 2080ti GPU.
The whole network is trained in an end-to-end manner using the ADAM optimizer~\cite{kingma2015adam}.
The learning rate is set to $0.0002$ and the batch size $N_{bs}$ is set as 16.
The image of each glyph is rendered with the font size 128 and resized into the resolution of 64 $\times$ 64.
The dimension of attribute embeddings $N_{e}$ is set as 64.
As we mentioned before, the category number of attributes $N_{\alpha}$ is set as 37.
The original attribute values in~\cite{o2014exploratory} vary from 0 to 100 and we re-scale them into [0,1] by dividing them by 100.
One training epoch consists of $\frac{N_{f} \cdot N_{c} }{N_{bs}} $ steps.
Empirically, we set $\lambda_{1} = 5$, $\lambda_{2} = 50$, $\lambda_{3} = 5$, $\lambda_{4} = 5$ and $\lambda_{5} = 20$ in loss functions.
\subsection{Effect of the Choice of Source Font}
\begin{figure}[t!]
  \centering
  \includegraphics[width=\columnwidth]{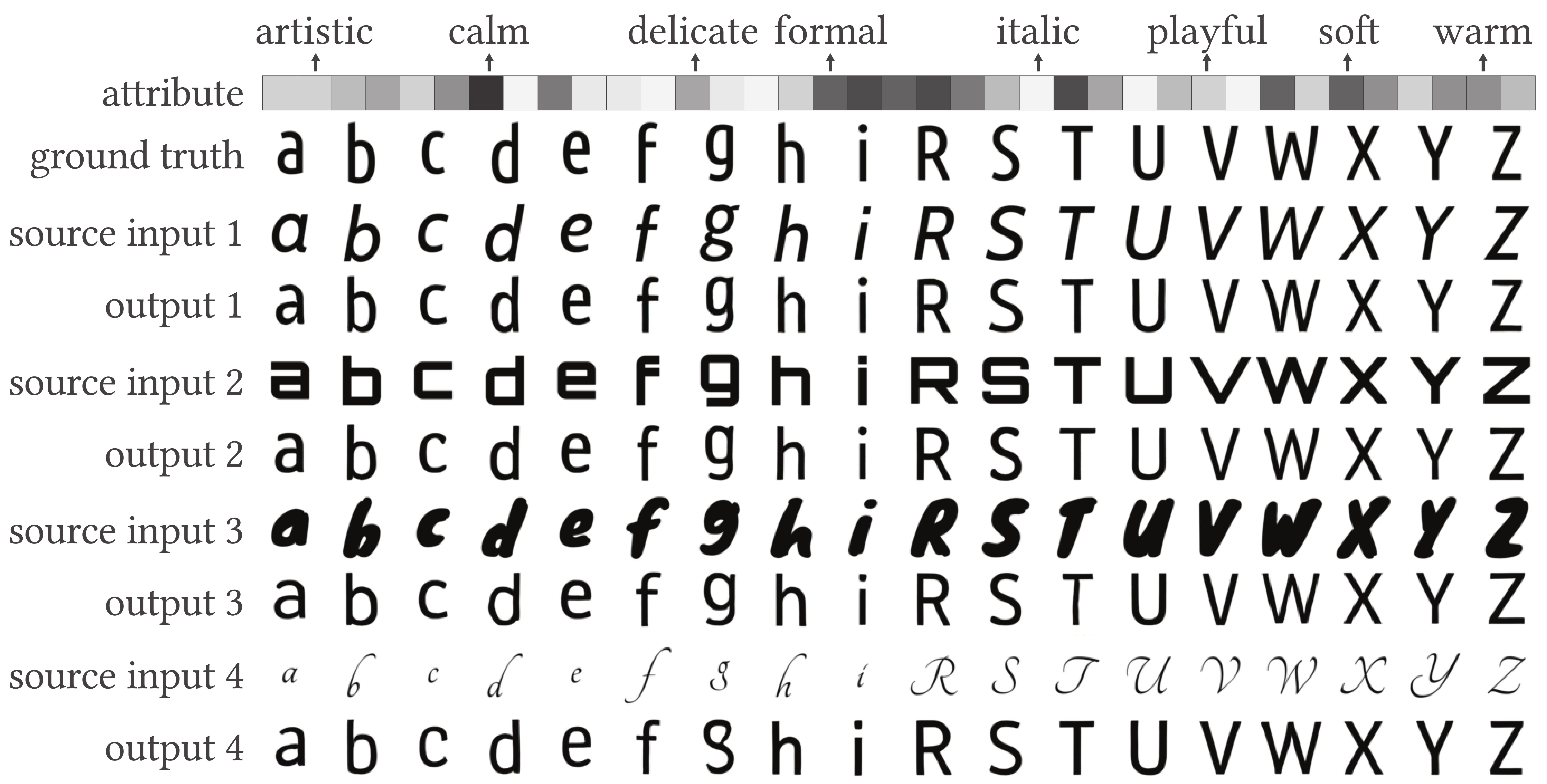}
  \caption{Generating glyph images with the same target attribute values from different source fonts. The pixel value of each grayscale grid represents each attribute's value. A darker grid indicates a higher attribute value.
  In the following figures we use the same way to display attribute values.}
  \label{fig:source_input}
\end{figure}

Intuitively, the font style of generated glyphs would be affected by the choice of source font in our model.
Surprisingly, through our experiments, we find that the influence is negligibly small when we select regular fonts as source.
In Fig.~\ref{fig:source_input}, we randomly select a font from the validation dataset as the target font and four different fonts from the training dataset as the source fonts.
The validation font is shown in the ``ground truth'' row and the four training fonts are shown in the ``source input 1'' - ``source input 4'' rows.
The generated glyph images from four different source fonts are shown in the ``output 1'' - ``output 4'' rows, respectively, with nearly the same shapes and subtle differences.
In the training process, the source font $a$ and the target font $b$ are both randomly selected and different fonts could match the same font as target.
This phenomenon suggests that our model manages to map different font styles to a fixed font style based on their attribute values.
We can also observe that the letter `g' transferred from the fourth font preserves the original artistic shape while the others are more regular.
In general, there is no strict restriction on the choice of source font but we highly recommend to choose a more regular font as the source.

\begin{figure*}[t!]
  \centering
  \includegraphics[width=\textwidth]{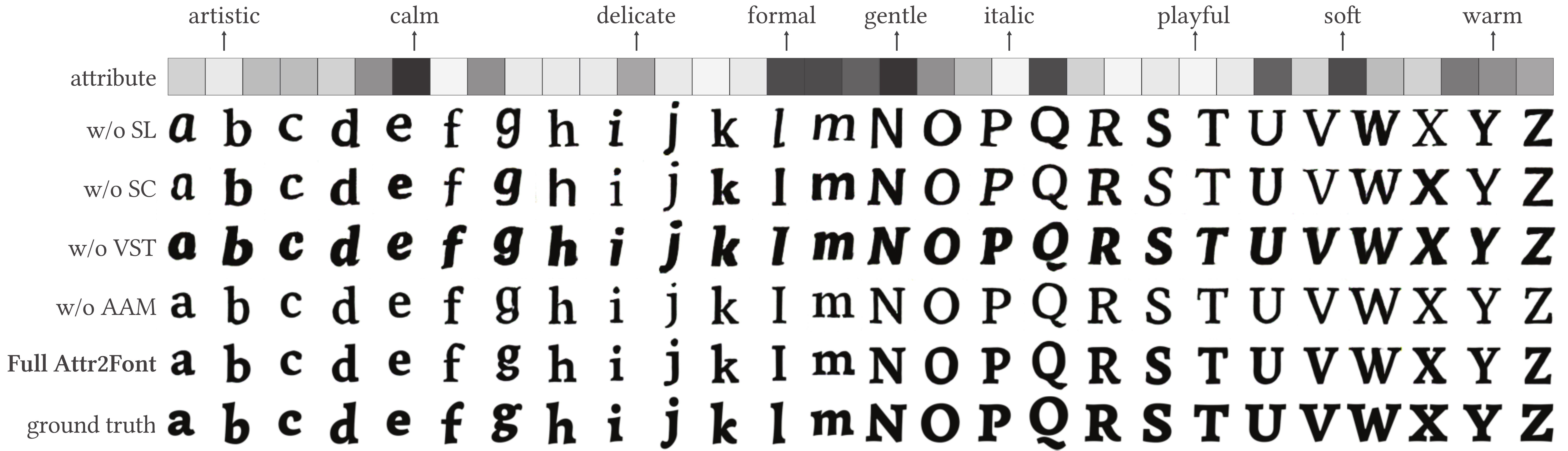}
  \caption{The glyph images generated by our models with different configurations.}
  \label{fig:ablation}
\end{figure*}

\subsection{Evaluation Metrics}
We adopt five commonly-used metrics for image generation to evaluate our model: Inception Score (IS)~\cite{salimans2016improved}, Fréchet Inception Distance (FID)~\cite{heusel2017gans}, Learned Perceptual Image Patch Similarity (LPIPS)~\cite{zhang2018unreasonable}, structural similarity index (SSIM) and pixel-level accuracy (pix-acc).
Specifically, IS is used to measure the realism and diversity of generated images.
FID is employed to measure the distance between two distributions of synthesized glyph images and ground-truth images, while SSIM aims to measure the structural similarity between them.
LPIPS evaluates the distance between image patches.
Since FID and IS cannot directly reflect the quality of synthesized glyph images, we also adopt the pixel-level accuracy (pix-acc) to evaluate performance.
Higher values of IS, SSIM and pix-acc are better, whereas for FID and LPIPS, the lower the better.
We also adopt two metrics which are widely used for shape matching, including the Hausdorff distance and Chamfer distance. We first extract the contours of glyph images using OpenCV and then utilize the points on the contours for calculating these two kinds of distances.

\subsection{Ablation Study}
For the purpose of analyzing the impacts of different
modules, we conduct a series of ablation studies by removing or changing the proposed modules in our model.
``Full Attribute2Font'' denotes the full proposed model.
“w/o skip-connection”, “w/o VST” and “w/o AAM”
denote the proposed models without the skip-connection, Visual Style Transformer and Attribute Attention Module, respectively.
``w/o unsupervised data'' denotes that we train our model only with the attribute-labeled fonts.

\subsubsection{Quantitative Experiments}
The loss curves of $l_{pixel}$, which can be found in the supplementary material, demonstrate how different modules promote the model's performance of reconstructing the target glyphs.
When finishing training, the quantitative results for ablation studies are shown in Table~\ref{tab:ablation}.
We witness a notable improvement brought by the proposed semi-supervised learning scheme via the exploitation of unlabeled fonts.
AAM also boosts our model's performance in a significant degree.
It is worth noting that higher IS score does not always guarantee higher quality of images.
The IS score is calculated from the classification predictions of a CNN trained with the ImageNet dataset. The domain mismatch between the glyph images and natural images makes the CNN consider the glyph images to be “unrealistic”.

\begin{table}
	\centering
	\caption{Quantitative results for ablation studies, w/o denotes without. ``SC'' denotes skip-connection. ``SL'' denotes semi-supervised learning.}
	\begin{tabular}{lccccc}
		\toprule
		              & IS      & FID      & LPIPS    & SSIM    & pix-acc   \\
		\midrule
		w/o SL        & \textbf{3.2654}  & 77.7443  & 0.11503  & 0.7098  & 0.7703     \\
		w/o SC        & 3.0067  & 38.5884  & 0.10981  & 0.7181  & 0.7919     \\
		w/o VST       & 3.1082  & 35.9498  & 0.09503  & 0.7366  & 0.8049     \\
		w/o AAM       & 3.0908  & 46.5920  & 0.08790  & 0.7502  & 0.7948     \\
		\textbf{Full Attr2Font} & 3.0740 & \textbf{26.8779} & \textbf{0.08742} & \textbf{0.7580} & \textbf{0.8153}  \\
		\bottomrule
	\end{tabular}

	\label{tab:ablation}
\end{table}

\subsubsection{Qualitative Experiments}
Fig.~\ref{fig:ablation} shows some examples of synthesized glyph images of our model under different configurations.
We select a representative font named as ``Simontta-Black'' from the validation dataset, as the target font for our model to generate.
Note that only the attribute values of ``Simontta-Black'' are sent into our model.
We expect our generated glyph images to share the most prominent styles with ``Simontta-Black'' instead of a perfect replica of ``Simontta-Black''.
The glyphs rendered with this font is shown in the ``ground truth" row.
The glyphs synthesized by our models with different configurations are shown in the second to sixth rows.
Without the semi-supervised learning or skip-connection or VST, our model tends to bring more artifacts on the synthesis results.
Without AMM, our model tends to miss some important characteristics such as ``delicate'' and ``warm''.
As we can see from Fig.~\ref{fig:ablation}, the effectiveness of each proposed module is vividly verified.

\subsection{Parameter Studies}
We conduct experiments to investigate how the number of glyph images sent into the style encoder (i.e., $m$) and the number of residual blocks in the Visual Style Transformer (i.e., $N_{rb}$), affect our model's performance.
The results are shown in Table~\ref{tab:parameter_m} and~\ref{tab:parameter_nrb}, from which we can observe that generally larger $m$ and $N_{rb}$ result in better performance.
However, larger values of $m$ and $N_{rb}$ increase the computational cost of our model.
To achieve a balance between the model size and performance, we choose $m$ = 4 and $N_{rb}$ = 16 as the default settings of our model unless otherwise specified.

\begin{table}
	\centering
	\caption{How the setting of $m$ affects our model's performance when $N_{rb}$ is fixed to 16.}
	\begin{tabular}{lccccc}
		\toprule
		Settings       & IS      & FID      & LPIPS    & SSIM    & pix-acc  \\
		\midrule
		$m$ = 1        & \textbf{3.1513}  & 56.8889  & 0.10355  & 0.7224  & 0.7836  \\
		$m$ = 2        & 3.1012  & 35.8145  & 0.09333  & 0.7423  & 0.8090  \\
		$m$ = 4        & 3.0740  & 26.8779  & 0.08742  & 0.7580  & 0.8153  \\
		$m$ = 8        & 3.0403  & 26.1309  & 0.08580  & 0.7592  & 0.8201  \\
		$m$ = 16       & 3.0907  & \textbf{25.5825}  & \textbf{0.07922}  & \textbf{0.7645}  & \textbf{0.8236}  \\
		\bottomrule
	\end{tabular}

	\label{tab:parameter_m}
\end{table}

\begin{table}
	\centering
	\caption{How the setting of $N_{rb}$ affects our model's performance when $m$ is fixed to 4.}
	\begin{tabular}{lccccc}
		\toprule
		Settings            & IS      & FID      & LPIPS    & SSIM    & pix-acc  \\
		\midrule
		$N_{rb}$ = 4        & 3.1720  & 29.0150  & \textbf{0.07494}  & 0.7338  & 0.7742   \\
		$N_{rb}$ = 16       & 3.0740  & 26.8779  & 0.08742  & 0.7580  & 0.8153   \\
		$N_{rb}$ = 32       & \textbf{3.1957}  & \textbf{24.0153}  & 0.07935  & \textbf{0.7862}  & \textbf{0.8308}   \\
		\bottomrule
	\end{tabular}

	\label{tab:parameter_nrb}
\end{table}

\begin{figure}
  \centering
  \includegraphics[width=\columnwidth,height=\textheight,keepaspectratio]{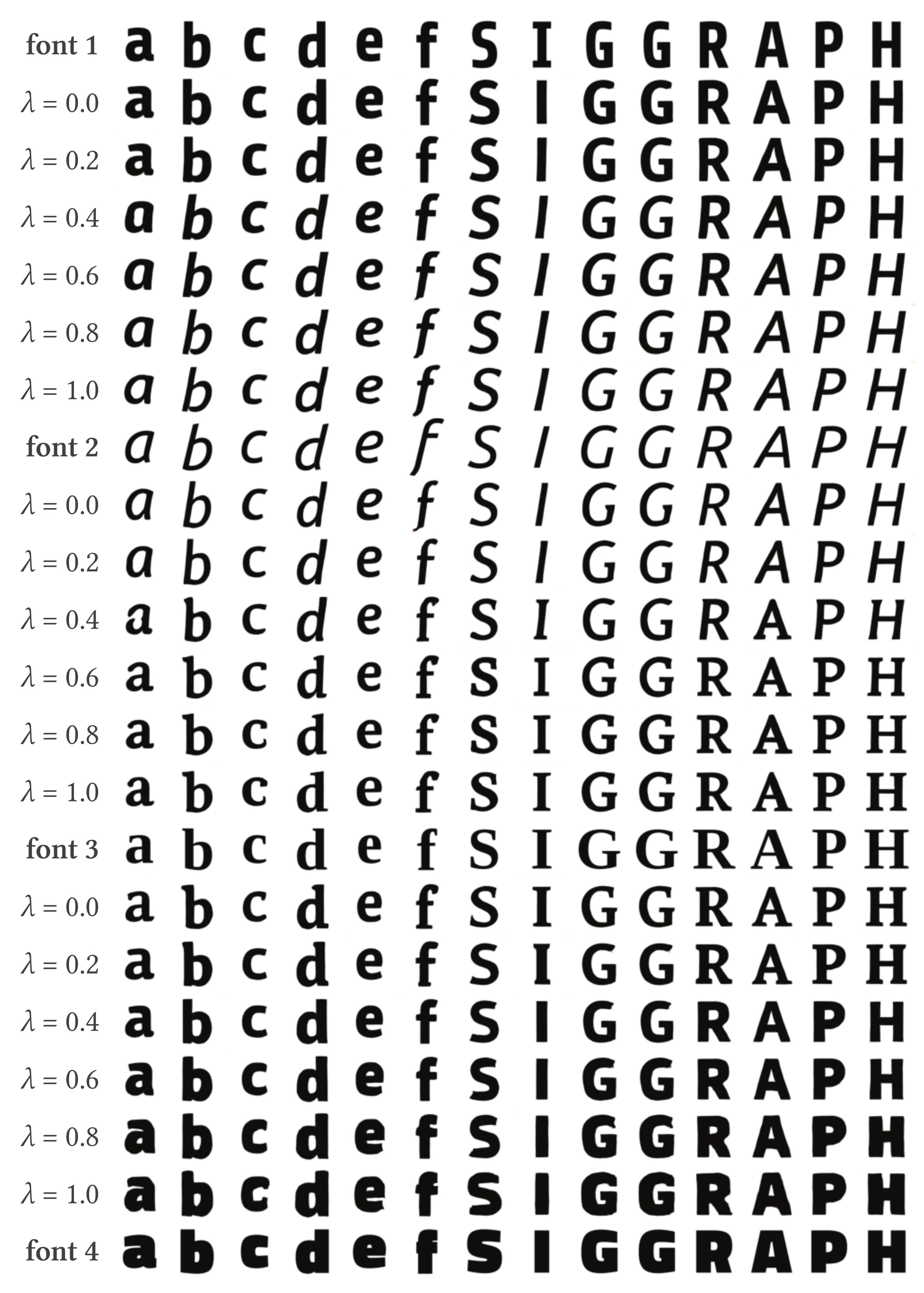}
  \caption{Generating glyph images by interpolation between the attribute values of two different fonts.
  Three interpolation processes (Font 1 to Font 2,  Font 2 to Font 3, Font 3 to Font 4) are presented in succession.}
  \label{fig:interp}
\end{figure}

\begin{figure*}[t!]
  \centering
  \includegraphics[width=\textwidth]{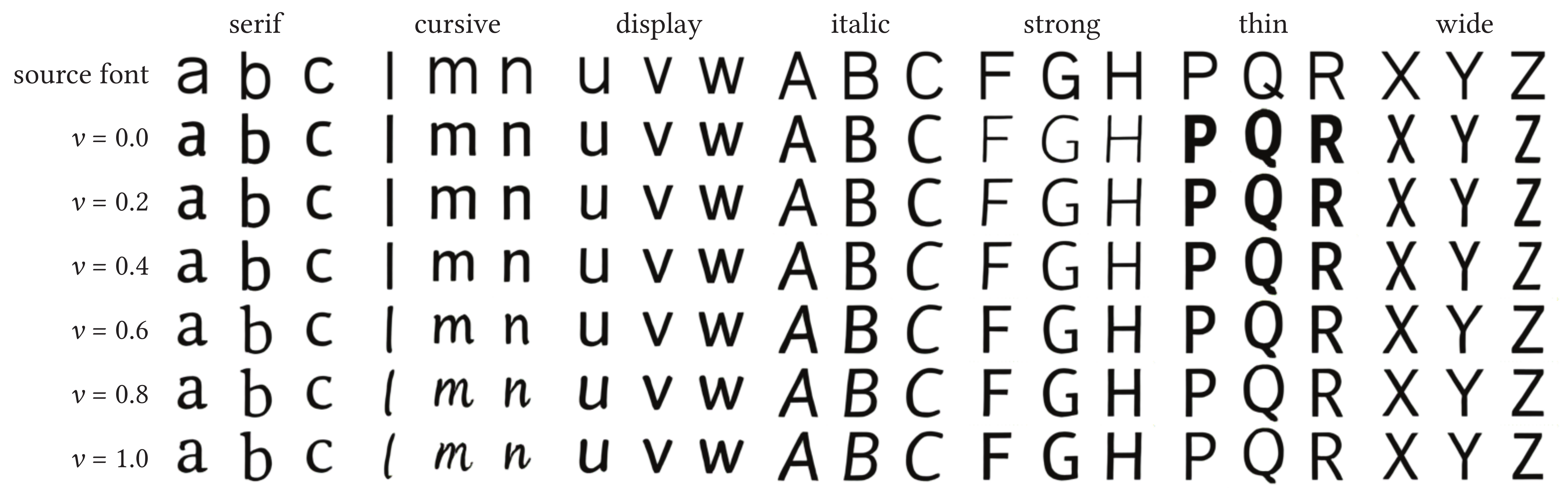}
  \caption{Editing fonts by changing the value of a single attribute.}
  \label{fig:editing}
\end{figure*}

\begin{figure}[t]
  \centering
  \includegraphics[width=\columnwidth]{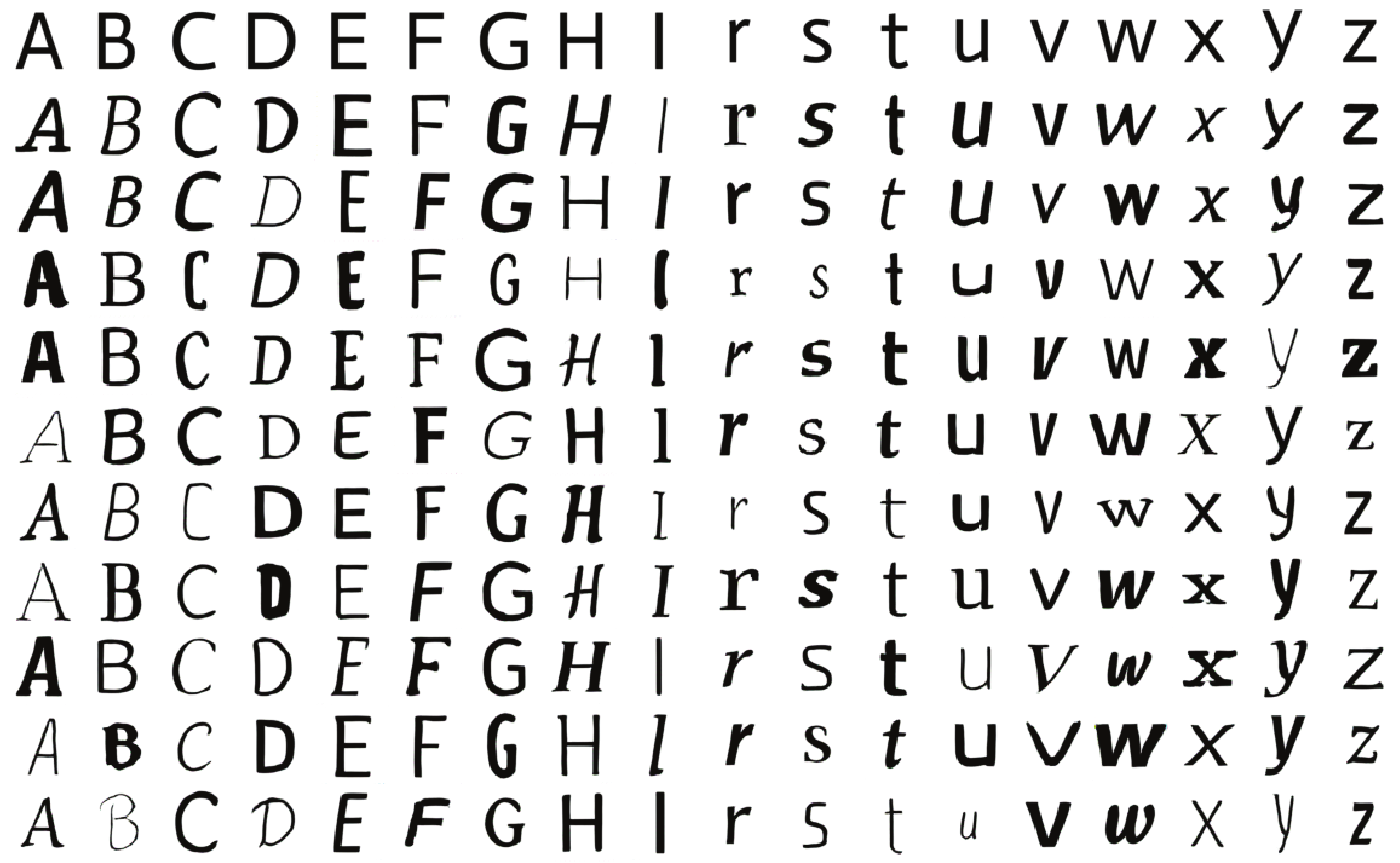}
  \caption{Generating glyphs from random attribute values.}
  \label{fig:random_target}
\end{figure}

\subsection{Attribute-controllable Interpolation}
Existing methods such as ~\cite{campbell2014learning} and ~\cite{guo2018creating} search for new fonts in an abstract manifold space.
Our model makes it feasible for interpolation between different fonts by explicitly modifying the font attribute values.
The interpolated attribute of two fonts is formulated as:
\begin{equation}
\alpha^{ip} = (1 -\lambda) \cdot \alpha(a) +  \lambda \cdot \alpha(b),
\end{equation}
where $\lambda \in [0,1]$ denotes the interpolation coefficient.
Fig.~\ref{fig:interp} shows that our model achieves smooth interpolation between different fonts and generates visually pleasing glyph images.
Compared to~\cite{campbell2014learning} and ~\cite{guo2018creating}, the synthesized glyph images from interpolated attribute values are more interpretable.

\subsection{Editing Fonts by Modifying Attribute Values}
\label{sec:edit_font}
Font designers often have the requirement of editing an existing font into their desired one by modifying just a few or single attributes.
In this section, we show our model's ability to delicately edit a font by manipulating the value of a single attribute.
We first set the font to be edited as the source font with attribute $\alpha(a)$.
Let $i^{*}$ be the index of the attribute category that we want to modify with the value of $v$, then we set the target attribute $\alpha(b)$ as:
\begin{equation}
\alpha_{i}(b)=\begin{cases}\alpha_{i}(a), & i \ne i^{*} \cr v, &i = i^{*}\end{cases},
\end{equation}
where $i$ is the index of attribute category and $\alpha_{i}$ is the value of the $i$-th attribute ($ 1 \le i \le N_{\alpha}$) .
Fig.~\ref{fig:editing} shows some examples of our synthesized glyph images by modifying the value of a single attribute, such as serif, cursive, display, italic, strong, thin and wide.
We set $v$ as 6 different values (0.0, 0.2, 0.4, 0.6, 0.8, 1.0) to illustrate the generated glyph images with 11 different levels of attribute values.
We can see that a smooth and gradual transition
can be achieved by our model when the attribute value varies from 0.0 to 1.0.

\subsection{Generating From Random Attribute Values}
Our model is capable of generating glyphs from random attribute values, regardless of the relationships among different attributes.
Fig.~\ref{fig:random_target} demonstrates the generated glyph images of our model from several random sets of attribute values.
The source font presented in the first row is fixed in this experiment and we randomly assign the attribute values of each character in the following rows (i.e., each character in the same row has different attribute values).
As we can see, the font styles of synthesized glyph images are highly varied and most of them are visually pleasing.
Our model can generate infinite typefaces from randomly assigned attribute values.
It is easy to find glyphs with high qualities and their corresponding sets of attribute values.
We believe this is inspiring for ordinary users and font designers to find or create their desired fonts.
We also demonstrate the glyph images of a whole char-set generated from random attribute values in the supplemental material.

\subsection{Attribute Study}
\begin{figure*}[t!]
  \centering
  \includegraphics[width=\textwidth]{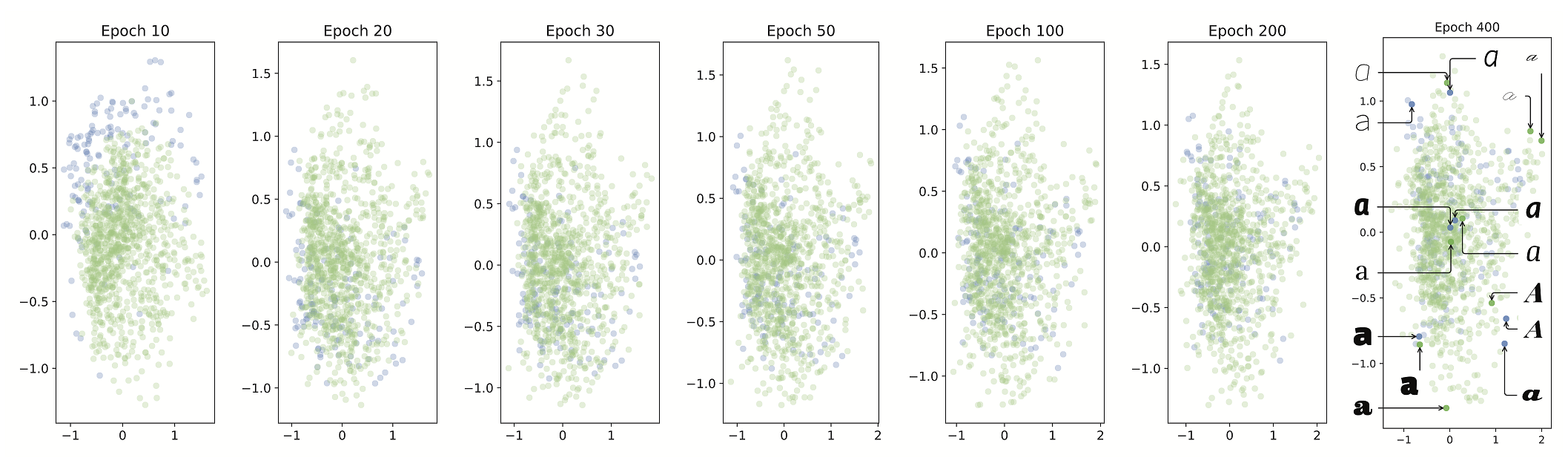}
  \caption{The distribution of attribute values varies with the training epochs. The green and blue points denote unlabeled and labeled fonts, respectively.}
  \label{fig:attribute_embedding_distribution}
\end{figure*}

\subsubsection{Distribution Analysis}
In this section, we illustrate how our semi-supervised learning scheme works by demonstrating the distribution of attribute values of different fonts (including both labeled fonts and unlabeled fonts) in Fig.~\ref{fig:attribute_embedding_distribution}.
The attribute values of each font are reduced into two dimensions by PCA (Principal Component Analysis).
The green points correspond to unlabeled fonts and the blue points correspond to fonts with labels.
As we can see from this figure, the distribution changes significantly in the early stage to fit the actual distribution of all font styles.
Afterwards, it remains relatively stable in the training phase.
In the beginning (Epoch 10), the green points are almost completely separated from the blue points.
In the later period, the green points are mixed into the blue points according to a specific distribution.
The attribute values of unlabeled fonts evolve from a stochastic state to a meaningful state, which verifies the effectiveness of our semi-supervised learning scheme.
We attach the glyph images (‘A’ or ‘a’) of some fonts to their corresponding points in Epoch 400.
We can observe that similar fonts are located closer while dissimilar fonts are located farther away.
Thereby, the predicted attributes of unlabeled fonts are reasonable.

\subsubsection{Impact of Different Attributes}
\label{sec:attr_impact}
\begin{figure}[t!]
  \centering
  \includegraphics[width=8cm]{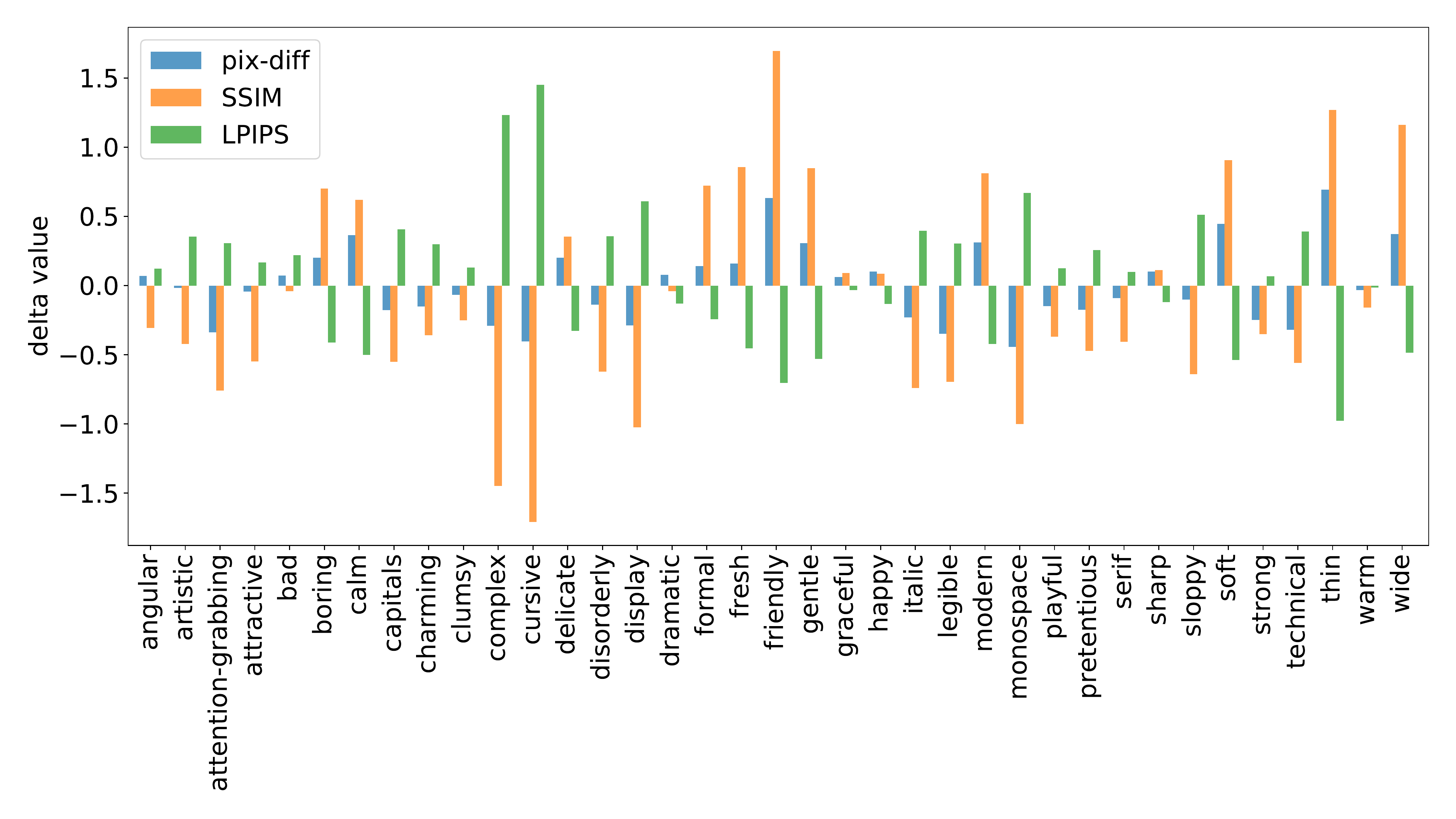}
  \caption{The impact of different attributes on the glyph shape. The horizontal axis represents all kinds of attributes in our model. The vertical axis represents the value change of three metrics (pix-diff, SSIM, LPIPS) after modifying each attribute's value. Please refer to Section~\ref{sec:attr_impact} for details.}
  \label{fig:influence_of_different_attributes}
\end{figure}

In this section, we investigate the impact of all kinds of attributes in font editing.
Quantitative statistics of the glyph changes are conducted after modifying the value of each target attribute.
The font editing is performed on all 28 fonts in the validation dataset.
When investigating the impact of the $i$-th attribute, the values of other attributes in $\alpha(b)$ are set to the same as $\alpha(a)$.
Let $\alpha_{i}(b)$ be the value of the $i$-th attribute of target font.
In the beginning we set $\alpha_{i}(b)$ to 0.0, and then increase $\alpha_{i}(b)$ by 0.2 each time until $\alpha_{i}(b)$ reaches 1.0.
We measure the shape difference of generated glyphs between the current step and the previous step by utilizing three metrics (pix-diff, SSIM, LPIPS), where
``pix-diff'' denotes the difference of pixel values.
We calculate the mean value of the five steps for three metrics respectively.
As we can see from Fig.~\ref{fig:influence_of_different_attributes}, nebulous attributes such as ``complex, cursive and friendly'' and concrete attributes such as ``thin and wide'' bring significant changes on the glyph shapes.

\subsubsection{Correlation between Attributes}

\begin{figure}[t!]
  \centering
  \includegraphics[width=\columnwidth]{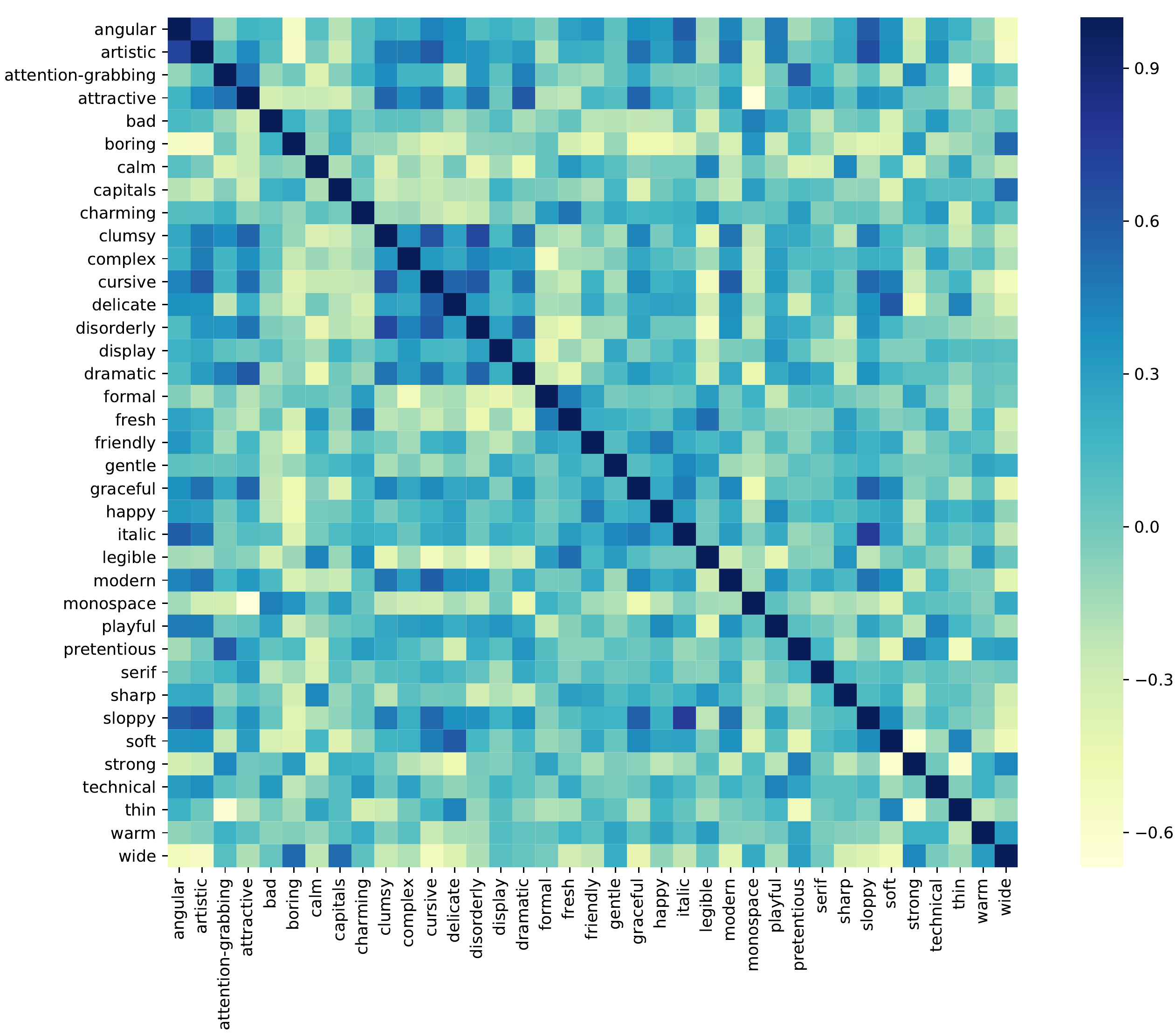}
  \caption{The correlation matrix of different attributes in the font dataset released by~\cite{o2014exploratory}.}
  \label{fig:atributes_correlation}
\end{figure}

In section~\ref{sec:method_des_ov}, we mention that it is unnecessary for our model to have attributes being independent of each other, making it very convenient for users to define their own attribute categories.
The only important thing is to accurately annotate the value of each attribute, instead of considering the correlation among different attributes.
To prove this statement, we investigate the correlation of different attributes in the font dataset employed in this paper.
The attributes are mutually related in a certain degree, which can be observed from Fig.~\ref{fig:atributes_correlation}.
This figure reveals some attribute pairs with strong correlations, such as ``sloppy'' and ``italic'', ``dramatic'' and ``attractive'', ``disorderly'' and ``clumsy''.
The correlation matrix also provides users with the guidance of manipulating attribute values into a meaningful set.

\subsection{Comparison with Other Models}
\subsubsection{Comparison with Attribute-controllable Image Synthesis Models}
\begin{figure}[t!]
  \centering
  \includegraphics[width=8cm]{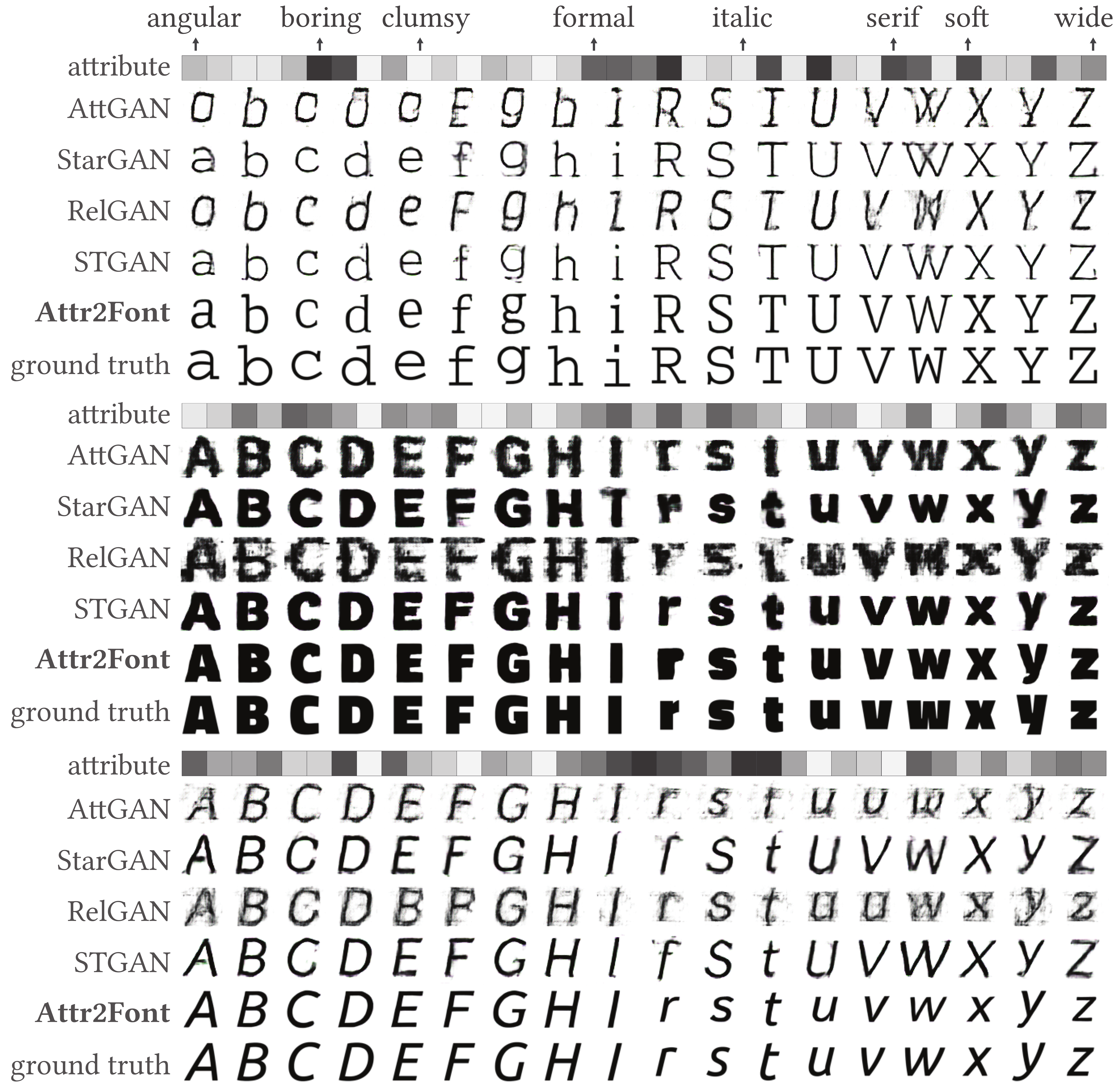}
  \caption{Comparison with existing methods of attribute-controllable image synthesis.}
  \label{fig:comparison}
\end{figure}

In this section, we compare our method with existing methods of attribute-controllable image synthesis, including AttGAN~\cite{he2019attgan}, StarGAN~\cite{choi2018stargan}, RelGAN~\cite{wu2019relgan} and STGAN~\cite{liu2019stgan}.
AttGAN and StarGAN are among the earliest works that address the problem of image attribute editing.
They tackle arbitrary attribute editing by taking the target attribute as input to the transform model.
STGAN and RelGAN utilize the difference between the target and source attributes as the transfer condition.
In Fig.~\ref{fig:comparison}, we select four fonts from the validation dataset whose attribute values are set as the target attributes in our model.
Because these existing GANs can only accept binary attribute values, the attribute values they receive are set to 0 or 1 if they are less or higher than 0.5.
Glyph images rendered from these fonts are shown in the ``ground truth" rows as reference.
We can see from Fig.~\ref{fig:comparison} that AttGAN and RelGAN tend to generate very blurry and low-quality glyphs.
STGAN generates glyphs with higher quality than AttGAN and RelGAN, which proves the effectiveness of the attribute difference.
Although the Selective Transfer Units (STU) were introduced to improve the skip-connection, STGAN still tends to bring many fake textures in the glyph images.
The training strategy of RelGAN makes it very unstable in our task and result in very bad results.
The quantitative results on the whole validation dataset are presented in Table~\ref{tab:comparison} which demonstrates that our model significantly outperforms the others.
We also compare the model size of different methods in Table~\ref{tab:comparison-para-prefer}, showing that our model possesses less parameters than STGAN but achieves much better performance than STGAN.

\begin{table}
	\centering
	\caption{Quantitative results of different image synthesizing models.}
	\resizebox{\linewidth}{!}{%
	\begin{tabular}{lccccccc}
		\toprule
		Model      & IS      & FID               & LPIPS    & SSIM   & pix-acc  & Hausdorff  & Chamfer       \\
		\midrule
		AttGAN     & 3.4008  & 200.1708          & 0.24039  & 0.6198  & 0.5287  & 11.2044  &   330.238 \\
		StarGAN    & \textbf{3.6179}  & 91.1436  & 0.12172  & 0.7024  & 0.7146  &  8.8748  &   317.818 \\
		RelGAN     & 3.1412  & 183.0307          & 0.23220  & 0.6216  & 0.5380  & 11.1048  &   339.822 \\
		STGAN      & 3.6178  & 83.3167           & 0.11779  & 0.7150  & 0.7444  &  8.7815  &   286.509 \\
		\textbf{Attr2Font} & 3.0740 & \textbf{26.8779} & \textbf{0.08742} & \textbf{0.7580} & \textbf{0.8153} &\textbf{7.1954} & \textbf{241.670} \\
		\bottomrule
	\end{tabular}}
	\label{tab:comparison}
\end{table}

\begin{table}
	\centering
	\caption{Comparison of model size and user preference for different image synthesizing methods.}
	\begin{tabular}{lcc}
		\toprule
		Model      & \# Parameters    & User prefer.    \\
		\midrule
		AttGAN     &          63.32M  & 0.0242   \\
		StarGAN    &  \textbf{53.33}M & 0.1020   \\
		RelGAN     &          61.74M  & 0.0305   \\
		STGAN      &          94.78M  & 0.1314   \\
		\textbf{Attr2Font}    & 69.85M & \textbf{0.7119}  \\
		\bottomrule
	\end{tabular}
    \label{tab:comparison-para-prefer}
\end{table}

\subsubsection{Comparison with Font Retrieval Models}

\begin{figure*}[t!]
  \centering
  \includegraphics[width=\textwidth]{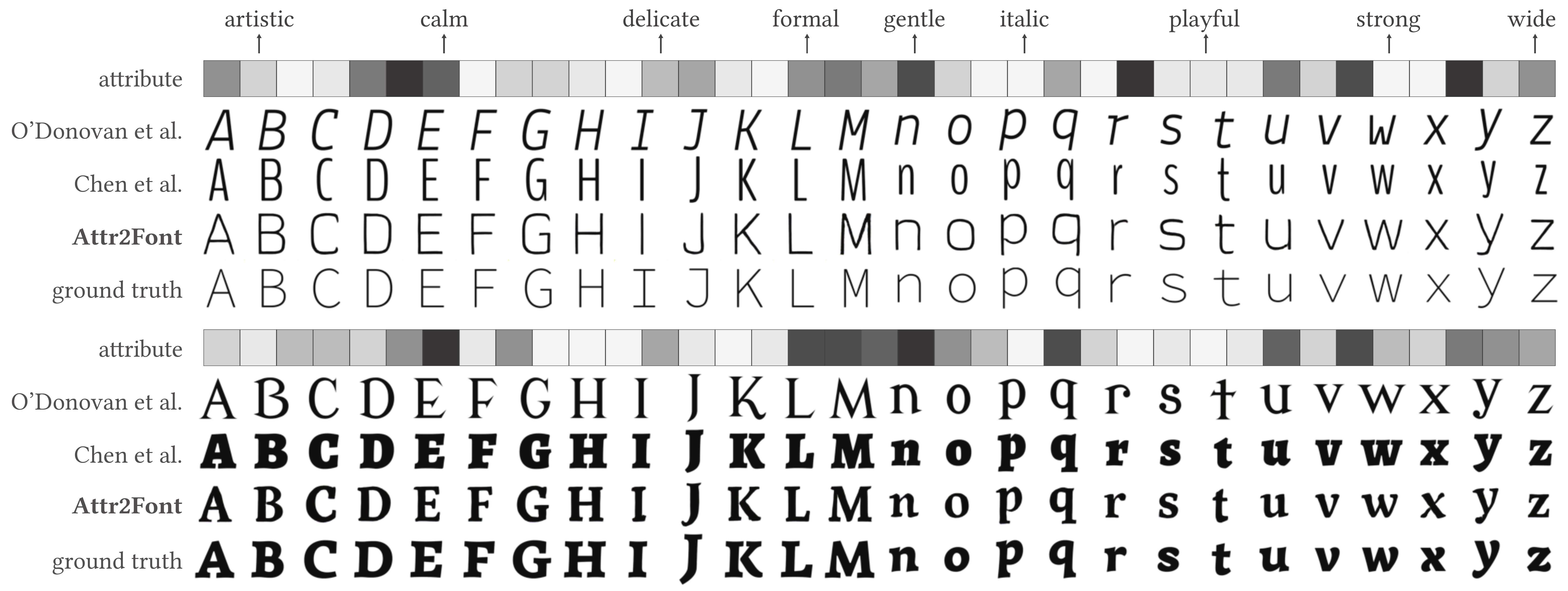}
  \caption{Comparison of our model and two existing font retrieval methods.}
  \label{fig:comparison_retrieval}
\end{figure*}

\begin{table}
	\centering
	\caption{Comparison of our model against two font retrieval methods.}
	\begin{tabular}{lcccc}
		\toprule
		Model             & Hausdorff  & Chamfer   & attr-acc  & User prefer.  \\
		\midrule
		O’Donovan et al.  & 10.2917    &  327.747  &  0.7735   & 0.0980        \\
		Chen et al.       &  7.9193    &  273.221  &  0.8213   & 0.2902        \\
		\textbf{Attr2Font}  &  \textbf{7.1954} &  \textbf{241.670} & \textbf{0.9372} & \textbf{0.6118}  \\
		\bottomrule
	\end{tabular}
	\label{tab:comparison-attr-acc}
\end{table}
Font retrieval models search in a font database and then return fonts which are most relevant to the attributes (or tags) in a query list.
However, choosing from existing fonts limits the model's flexibility and the variety of available fonts.
The lack of flexibility leads the retrieved fonts to miss some important attribute characteristics.
For example, our model achieves smooth and continuous interpolation between different fonts.
But when we set the interpolated attributes of two different fonts as query, the font retrieval models cannot provide such diverse and precise results on account of the limited number of existing fonts.
Increasing the scale and diversity of font database may solve the problem to some extent but the computational cost will also increase dramatically.
Another problem with the font retrieval models is the ignorance of some remarkable attributes in pursuit of global similarities.
Fig.~\ref{fig:comparison_retrieval} compares the synthesis results of our model with the retrieval results of~\cite{o2014exploratory} and~\cite{chen2019large} when given two sets of attribute values in the validation dataset.
Our model obtains the most precise result compared to the other two methods.
In the first example, \cite{o2014exploratory} ignores the weak ``italic'' attribute and \cite{chen2019large} ignores the strong ``wide'' attribute.
In the second example, \cite{o2014exploratory} ignores the strong ``formal'' attribute and \cite{chen2019large} ignores the weak ``strong'' attribute.
We train an attribute predictor and implement it on glyph images outputted from our model and these two font retrieval methods, respectively.
As shown in Table~\ref{tab:comparison-attr-acc}, our model achieves the highest accuracy, which gives a solid evidence of our method's superiority to the state of the art.
\subsubsection{User Study}
We conduct a user study among ordinary users to compare our model with other existing methods.
The first part of this study investigates the users' preference among AttGAN, StarGAN, RelGAN, STGAN and our model.
Specifically, for each font in the validation dataset, we send the attribute values of this font into the five models respectively, and get five sets of generated glyph images.
For each set of glyph images, a participant is asked to choose the one that possesses the best quality and has the most similar style as the glyph images rendered by the validation font.
The second part of this study investigates the users' preference among two above-mentioned font retrieval methods and our model.
For each set of provided glyph images, a participant is asked to choose the one that corresponds best to the given attribute values.
50 participants have taken part in this user study.
Participants in this user study consist of students and faculties from different colleges.
Most of them have no expertise in font design.
Statistical results are shown in the ``User prefer.'' column of Table~\ref{tab:comparison-para-prefer} and~\ref{tab:comparison-attr-acc}, respectively, from which we can see that our model outperforms the others by a large margin. \par

\begin{figure}[t!]
  \centering
  \includegraphics[width=8cm]{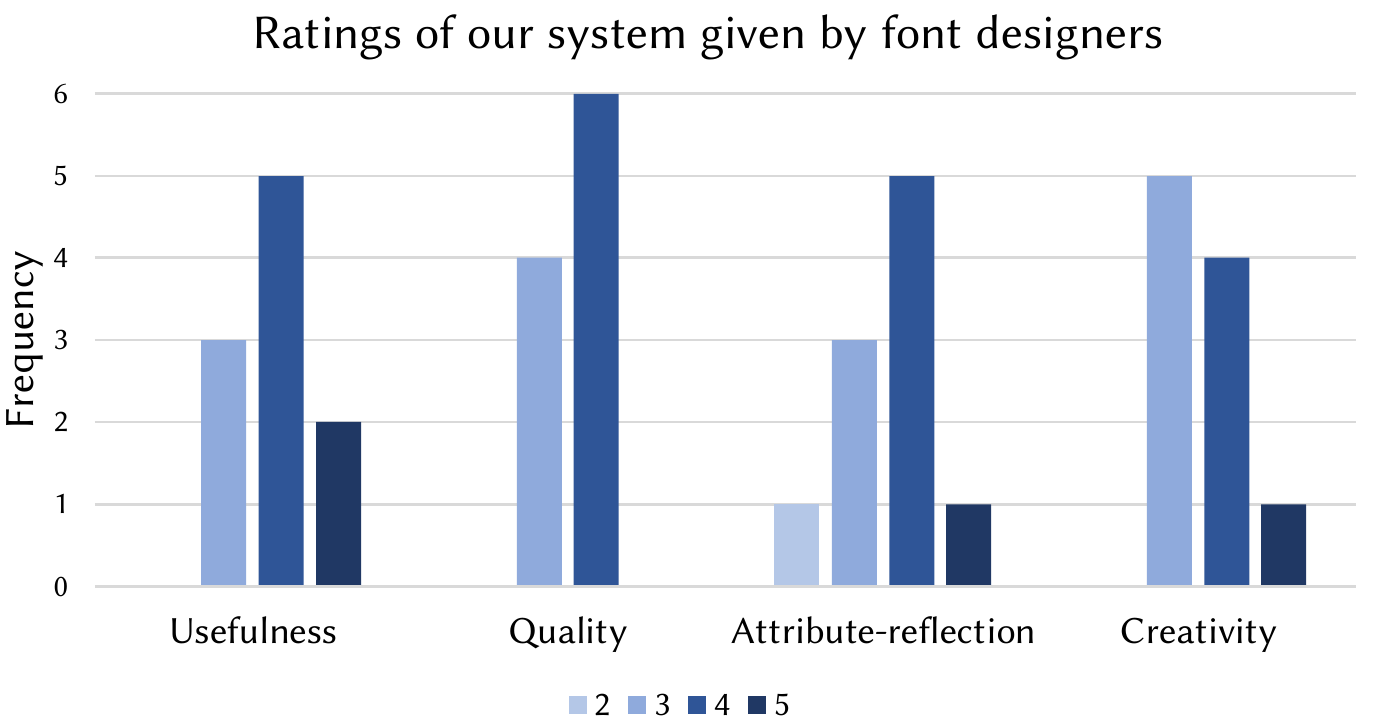}
  \caption{A user study among professional font designers. 10 participants have evaluated our model from 4 different perspectives and give ratings which are among 1-5 (higher is better).}
  \label{fig:user_study_designers}
\end{figure}

Our model is initially designed for ordinary users, but it can also inspire/help professional designers to create new vector fonts based on the synthesized glyph images.
There exists a widely-used font designing procedure that is to first obtain characters’ raster images and then automatically/manually convert them to vector glyphs.
To further verify whether our system is helpful in practical use, we conduct another user study among professional font designers in Founder Group, one of the world's largest font producers.
10 professional font designers have taken part in this user study.
We develop a user interface for them and they are allowed to arbitrarily manipulate the attribute values and observe our generated glyph images.
After a deep experience, they are asked to answer the following questions:
(1) Is the system useful for font designers? (2) Does the system provide glyph images which are visually pleasing and embody the input attributes? (3) Can some of those machine-generated fonts inspire their creations?
We conclude the 4 different perspectives as usefulness, quality, attribute-reflection and creativity, respectively.
For each perspective, they give a rating among 1 to 5, where 1 denotes the most negative and 5 denotes the most positive.
The results are shown in Fig.~\ref{fig:user_study_designers}, from which we can see that most of the designers affirm the practical usefulness of our system and agree that it can generate satisfactory glyph images.
Many of them have been deeply inspired by some creative fonts synthesized by our system and are willing to convert them into vector fonts.
For instance, a senior font designer spoke highly of our system by commenting : ``it is very useful for assisting and inspiring font designers and will significantly improve the efficiency of font design and production in the future."

\subsection{Application on Chinese Fonts}
\begin{figure}[t!]
  \centering
  \includegraphics[width=8cm]{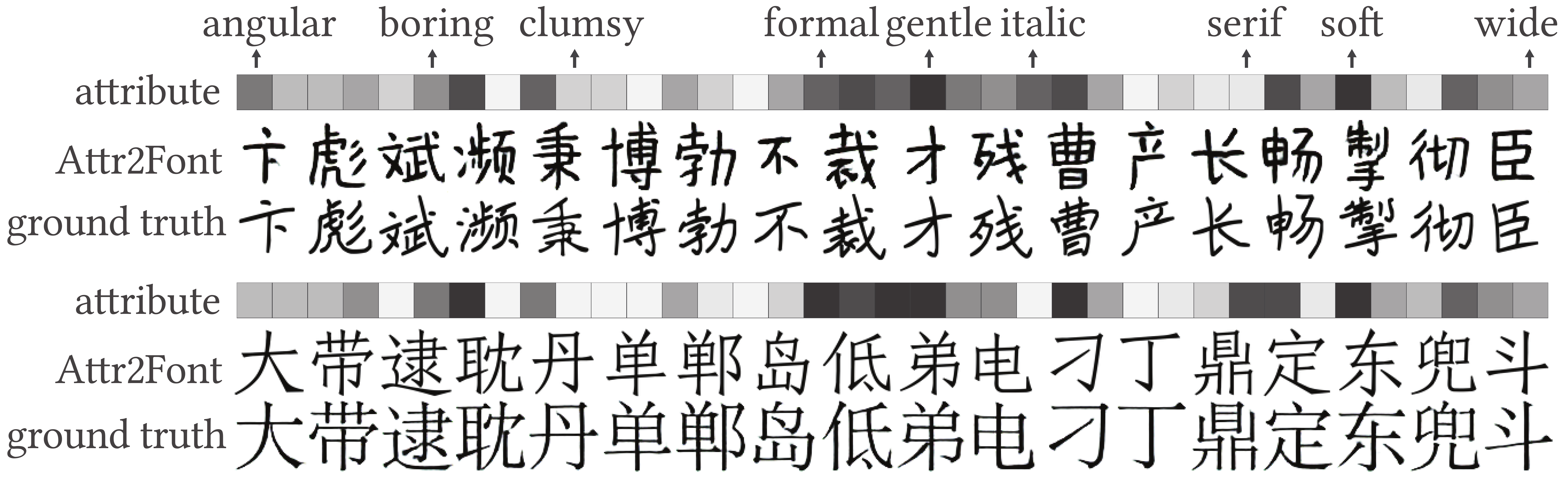}
  \caption{Generating Chinese fonts from attributes.}
  \label{fig:cn_synthesis}
\end{figure}
\begin{figure}[t!]
  \centering
  \includegraphics[width=8cm]{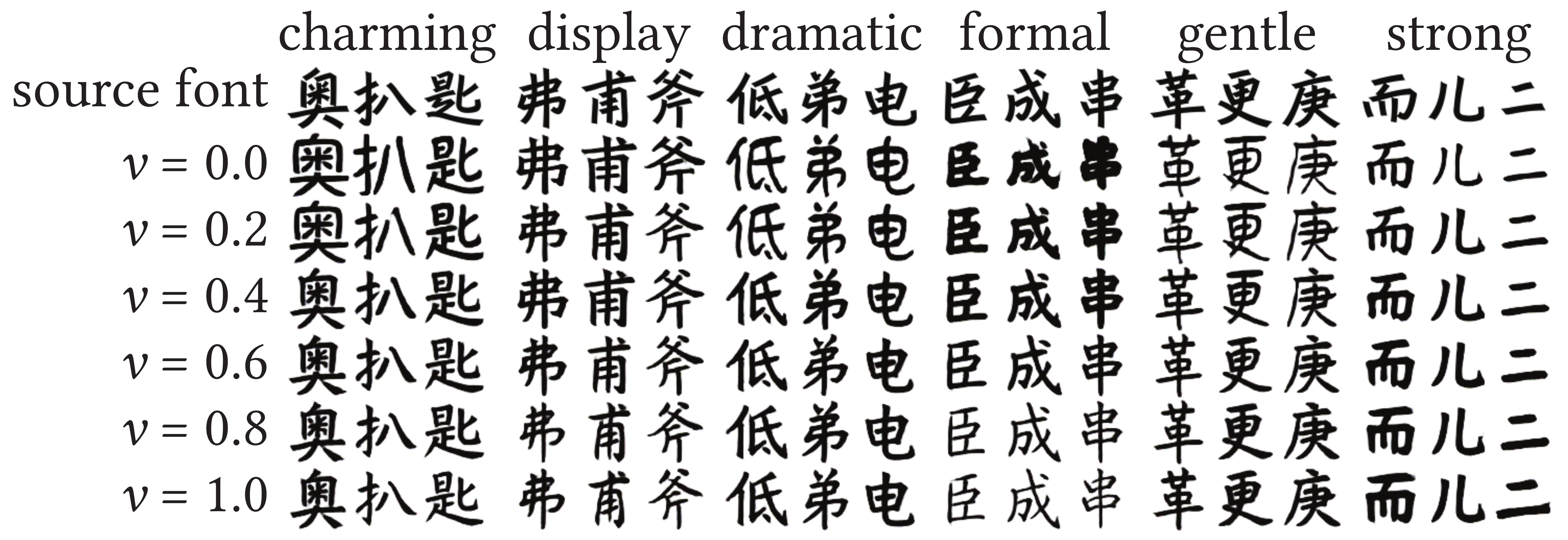}
  \caption{Editing Chinese fonts by changing a single attribute's value.}
  \label{fig:cn_editing}
\end{figure}
\begin{figure}[t!]
  \centering
  \includegraphics[width=8cm]{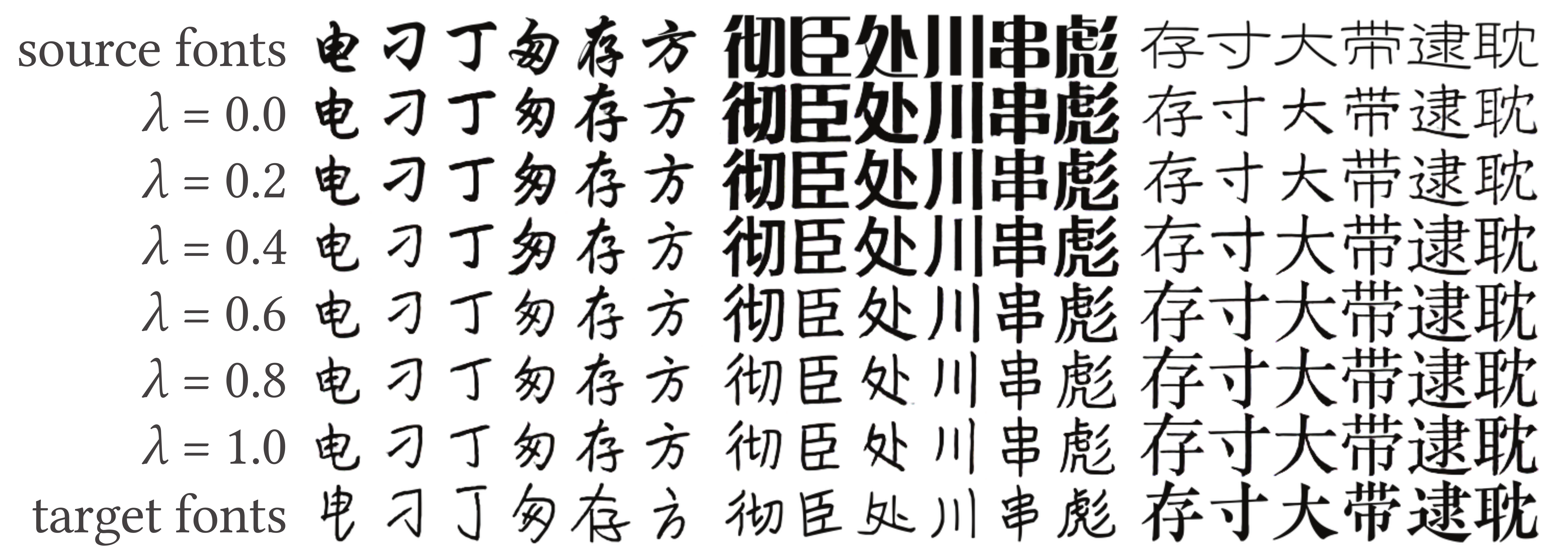}
  \caption{Interpolation among Chinese fonts.}
  \label{fig:cn_interp}
\end{figure}

To verify the generality of our model, we apply our model on Chinese fonts in addition to English fonts.
We collected 245 Chinese fonts to form a new dataset named as AttrFont-CN.
These fonts contain both Chinese and English glyphs and in general the Chinese and English glyphs in the same font share the same style.
78 fonts from AttrFont-ENG and 78 fonts from AttrFont-CN are matched in terms of their font style similarity on English glyphs.
Then we annotate the latter with the attribute values of the former.
The selected 78 fonts in AttrFont-CN are divided into 50 training fonts and 28 validation fonts.
There are altogether 217 training fonts, including 50 labeled fonts and 167 unlabeled fonts, and 28 validation fonts.
We train our model with Chinese glyphs from a character set with the size of 100.
In Fig.~\ref{fig:cn_synthesis},~\ref{fig:cn_editing} and~\ref{fig:cn_interp} we demonstrate our model's performance on synthesizing glyph images from attributes, editing fonts and interpolation among fonts, respectively.
Our model still achieves good performance on creating Chinese fonts from attributes although Chinese glyphs are much more complicated than English glyphs.

\section{Discussion}
\subsection{Generating Fonts in Strange Styles}
\begin{figure}[t!]
  \centering
  \includegraphics[width=8cm]{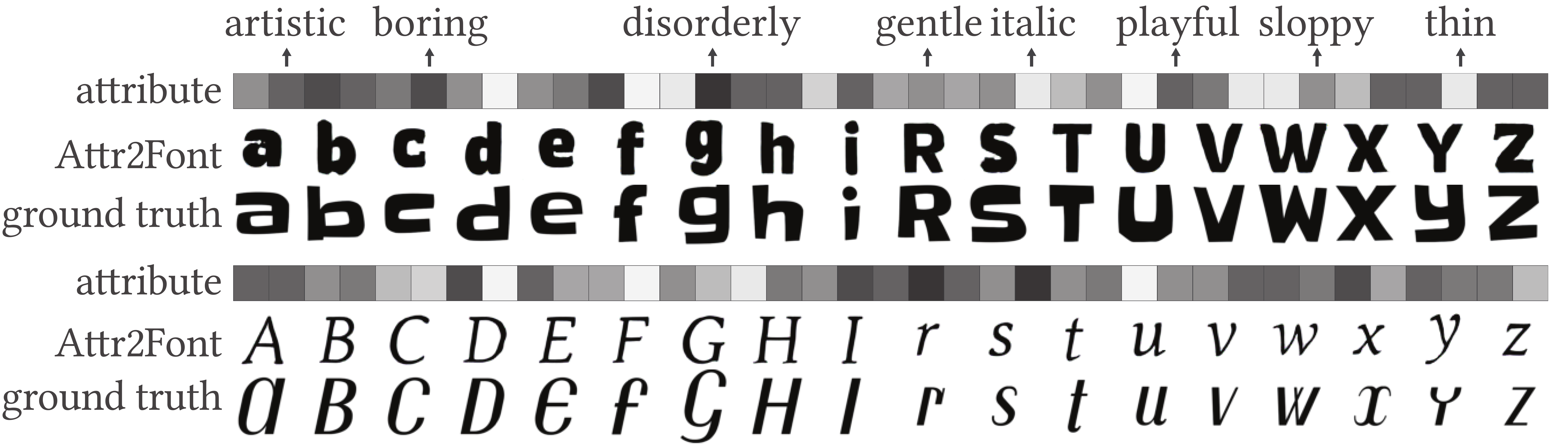}
  \caption{Generating fonts in strange styles.}
  \label{fig:difficult_target}
\end{figure}

Fig.~\ref{fig:difficult_target} shows some synthesis results of our model in strange font styles.
The two target fonts are selected from the validation dataset and possess very unique styles.
Our generated results share some remarkable styles with the ground truth although there exist some differences.

\subsection{Limitations}
\begin{figure}[t!]
  \centering
  \includegraphics[width=8cm]{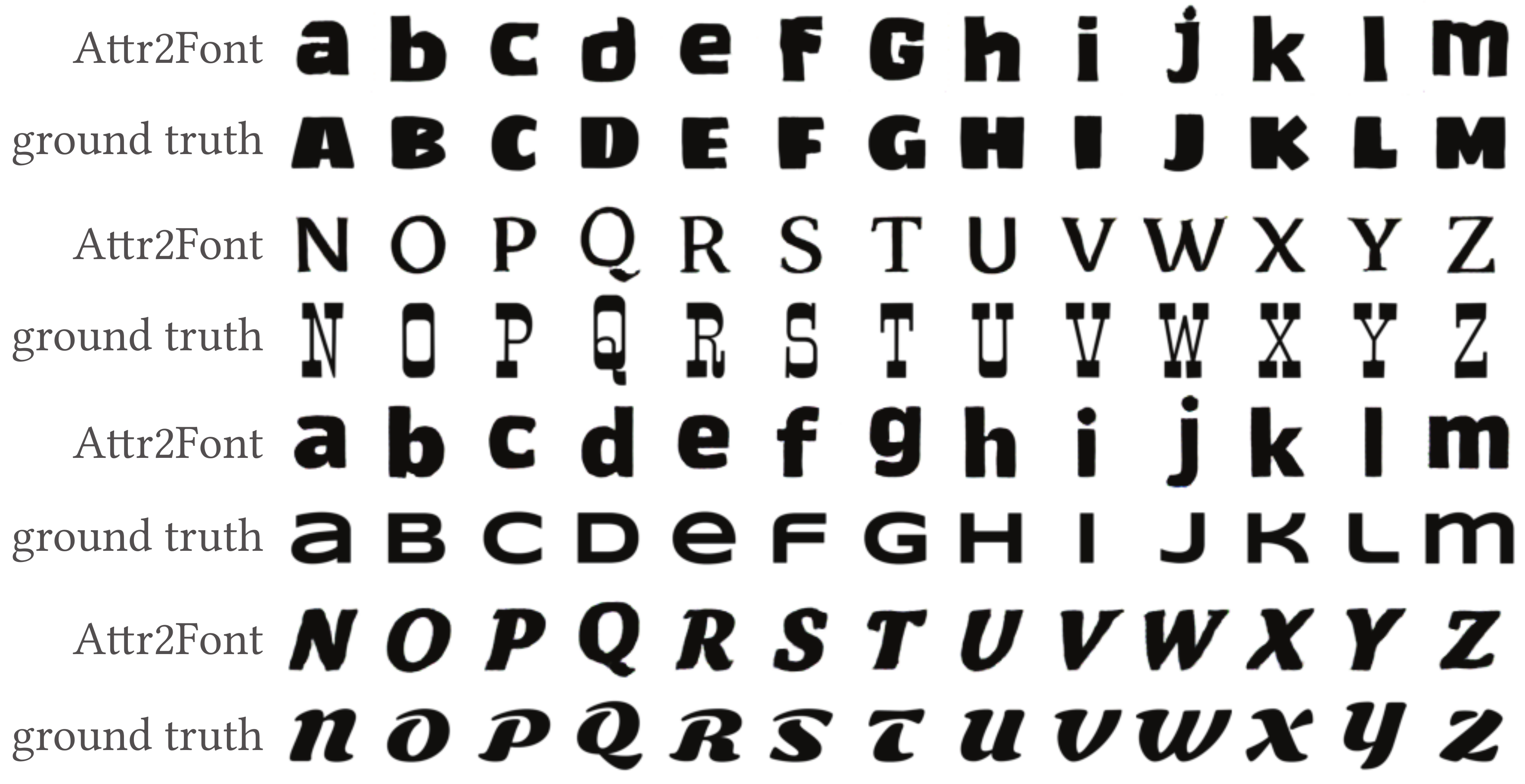}
  \caption{Some failure cases of our model.}
  \label{fig:limitation}
\end{figure}
Fig.~\ref{fig:limitation} reveals two limitations of our model:
(1) The encoder-decoder architecture usually requires that the source image and the target image should share a similar geometric structure.
The upper and lower cases of some characters have very different shapes.
Thereby, the proposed model cannot handle capital fonts well, producing lowercase glyph images instead of capital ones even though the attribute ``capital'' is specified.
Besides, `a' and `g' both have two different typologies. If we transfer one typology to another, the generated glyph tends to preserve the structure of the source typology to some extend (see the last row in Fig.~\ref{fig:source_input}).
(2) The assumption of each font corresponds to a set of font attribute values is not completely accurate.
A novel font may have some other characteristics beyond the pre-defined attributes.
If a desired font style is so unique that the pre-defined attributes cannot sufficiently describe it, our model will certainly fail to portray the font style, such as the second to fourth cases in Fig.~\ref{fig:limitation}.

\subsection{Application in Other Image Synthesis Tasks}
\label{sec:app_on_other_tasks}
\begin{figure}[t!]
  \centering
   \includegraphics[width=\columnwidth]{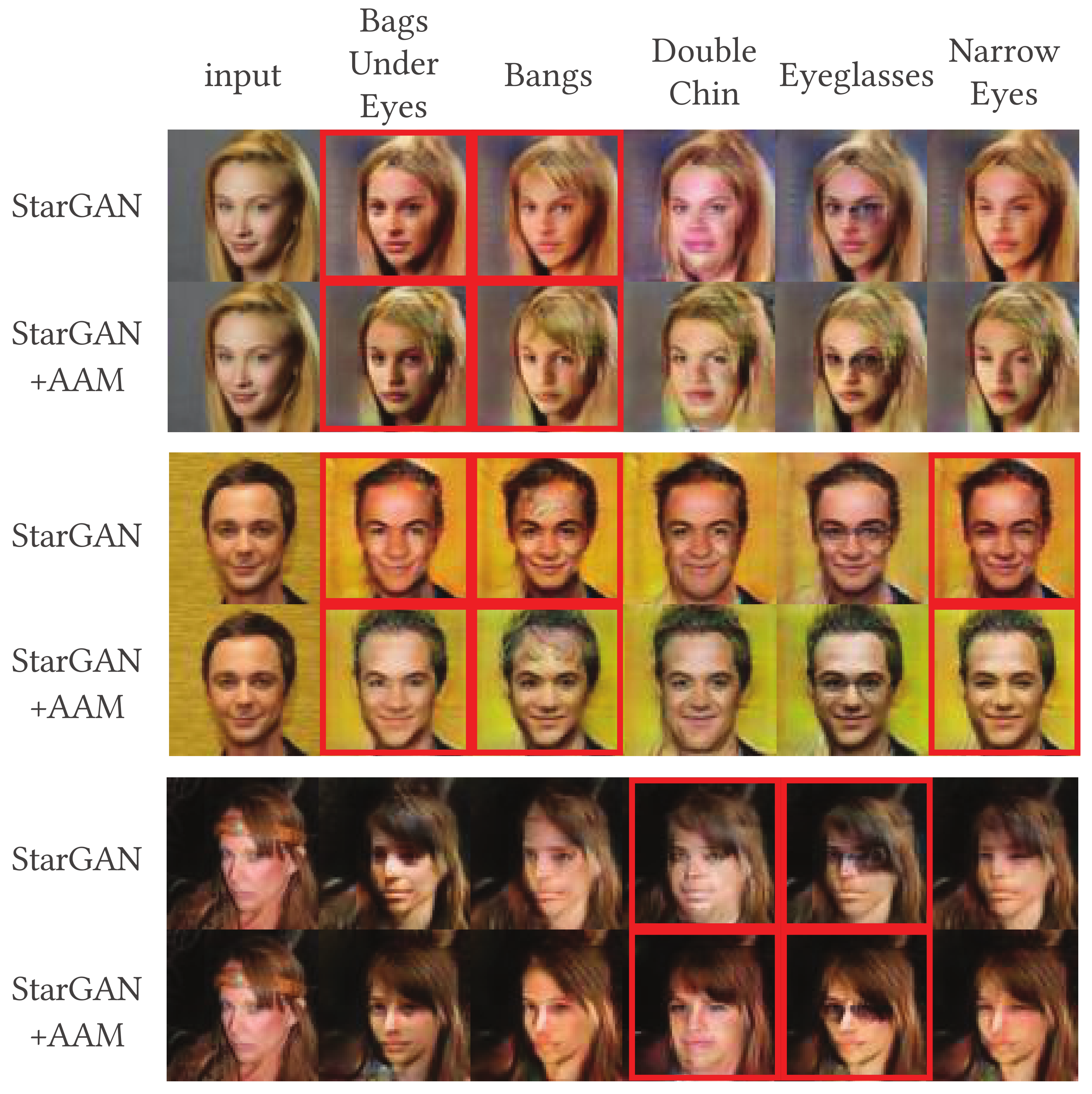}
   \caption{Comparison of StarGAN+AAM and StarGAN. The red rectangles emphasize some cases where StarGAN+AAM significantly differs with StarGAN. Note that StarGAN suffers from the problem of identity inconsistency. Thereby, synthesized faces may look quite different against input faces in this figure.}
   \label{fig:stargan_aam}
 \end{figure}

To further verify the generality of our Attribute Attention Module (AAM), we apply it on the task of face image synthesis.
We integrate AAM into StarGAN (denoted as StarGAN + AAM) and compare it with the original StarGAN.
Specifically, AAM is imposed on the different layers of the decoder in StarGAN, which is the same as our model.
We show some cases in Fig.~\ref{fig:stargan_aam} to compare the performance of these two models, where the input images are selected from a publicly-available database~\cite{liu2015faceattributes} whose training set is adopted to train the models.
More results are presented in the supplemental material.\par
Generally, our model aims to handle the task of image-to-image translation conditioned on attributes.
Thereby, potentially our model can also be applied in many other tasks, such as scene translation according to timing (from day to night), climate (from spring to winter), etc.
Theoretically, the semi-supervised learning scheme is applicable to any scenarios when there is a shortage of attribute annotations.

\subsection{Exploring the Variations of Our Work}
To sum up, our work learns a style mapping function based on the attribute difference.
Previous to this paper, style similarity metrics have been widely used in design systems, such as~\cite{o2014exploratory,garces2014similarity,lun2015elements,serrano2018intuitive}.
The goals of both the style similarity metrics and our mapping function are to build relationships between different styles.
However, our mapping function concentrates on mapping one style to another according to the attribute difference while those existing style similarity metrics aim to calculate affinity scores between different styles.
As a result, our system performs well for the task of synthesizing novel styles by arbitrarily assigning attribute values while those metric-learning systems are good at matching existing styles for specific needs.
These two kinds of systems can benefit from each other:
(1) The synthesized styles and existing styles can be combined to increase the diversity of styles and better meet the design needs.
(2) The feature representations of these two kinds of systems might be complementary.
In metric-learning systems, features of different styles are typically extracted/learnt individually.
In our system, features of different styles are mutually learnt from the mappings between them.
(3) Our semi-supervised learning scheme can also be applied to measure the style similarity when the manually annotated data are insufficient for supervised learning.
\par
We also come up with some other potential variations of our work.
A simple variation of our work is to synthesize some other kinds of objects (other 2D images and 3D shapes, etc.), which has been partially discussed in Section~\ref{sec:app_on_other_tasks}.
Previous works, such as StarGAN, STGAN, etc., are limited to synthesizing faces or fashions from discrete attribute values, but our methods go one step further.
Another variation of our work is to incorporate controllable attributes into the task of style transfer, which aims to change the style of a content image according to a reference style image.
In previous works, such as~\cite{gatys2016image} and~\cite{ma2014analogy}, the styles of the reference image are fully blended into the content image.
With the help of our methods, the reference image’s styles can be selectively transferred through the control of attributes.
\section{Conclusion}
In this paper, we presented a novel model to create new fonts by generating glyph images in accordance with user-specified attributes and their corresponding values.
The Attribute Attention Module and semi-supervised learning scheme were introduced to deal with the difficulties in this task.
Extensive experiments were conducted to demonstrate the effectiveness and superiority of our model compared to existing approaches.
Last but not least, our model is not only limited to generating English glyphs, but also applicable to synthesizing fonts in any other writing systems (e.g., Chinese).
We would like to investigate the following issues in the future: (1) extending our method to directly handle the task of vector font generation; (2) constructing a 2D font manifold for users to easily explore the fonts generated by our system; (3) improving the architecture of deep learners to promote the quality of generated fonts.

\begin{acks}
This work was supported by National Natural Science Foundation of China (Grant No.: 61672043 and 61672056), Beijing Nova Program of Science and Technology (Grant No.: Z191100001119077), Center For Chinese Font Design and Research, and Key Laboratory of Science, Technology and Standard in Press Industry (Key Laboratory of Intelligent Press Media Technology).
\end{acks}

\bibliographystyle{ACM-Reference-Format}
\bibliography{sample-bibliography}


\begin{thebibliography}{36}


\ifx \showCODEN    \undefined \def \showCODEN     #1{\unskip}     \fi
\ifx \showDOI      \undefined \def \showDOI       #1{#1}\fi
\ifx \showISBNx    \undefined \def \showISBNx     #1{\unskip}     \fi
\ifx \showISBNxiii \undefined \def \showISBNxiii  #1{\unskip}     \fi
\ifx \showISSN     \undefined \def \showISSN      #1{\unskip}     \fi
\ifx \showLCCN     \undefined \def \showLCCN      #1{\unskip}     \fi
\ifx \shownote     \undefined \def \shownote      #1{#1}          \fi
\ifx \showarticletitle \undefined \def \showarticletitle #1{#1}   \fi
\ifx \showURL      \undefined \def \showURL       {\relax}        \fi
\providecommand\bibfield[2]{#2}
\providecommand\bibinfo[2]{#2}
\providecommand\natexlab[1]{#1}
\providecommand\showeprint[2][]{arXiv:#2}

\bibitem[\protect\citeauthoryear{Azadi, Fisher, Kim, Wang, Shechtman, and
  Darrell}{Azadi et~al\mbox{.}}{2018}]%
        {azadi2018multi}
\bibfield{author}{\bibinfo{person}{Samaneh Azadi}, \bibinfo{person}{Matthew
  Fisher}, \bibinfo{person}{Vladimir~G Kim}, \bibinfo{person}{Zhaowen Wang},
  \bibinfo{person}{Eli Shechtman}, {and} \bibinfo{person}{Trevor Darrell}.}
  \bibinfo{year}{2018}\natexlab{}.
\newblock \showarticletitle{Multi-content gan for few-shot font style
  transfer}. In \bibinfo{booktitle}{\emph{Proceedings of the IEEE conference on
  computer vision and pattern recognition}}. \bibinfo{pages}{7564--7573}.
\newblock


\bibitem[\protect\citeauthoryear{Balashova, Bermano, Kim, DiVerdi, Hertzmann,
  and Funkhouser}{Balashova et~al\mbox{.}}{2019}]%
        {balashova2019learning}
\bibfield{author}{\bibinfo{person}{Elena Balashova}, \bibinfo{person}{Amit~H
  Bermano}, \bibinfo{person}{Vladimir~G Kim}, \bibinfo{person}{Stephen
  DiVerdi}, \bibinfo{person}{Aaron Hertzmann}, {and} \bibinfo{person}{Thomas
  Funkhouser}.} \bibinfo{year}{2019}\natexlab{}.
\newblock \showarticletitle{Learning A Stroke-Based Representation for Fonts}.
  In \bibinfo{booktitle}{\emph{Computer Graphics Forum}},
  Vol.~\bibinfo{volume}{38}. Wiley Online Library, \bibinfo{pages}{429--442}.
\newblock


\bibitem[\protect\citeauthoryear{Campbell and Kautz}{Campbell and
  Kautz}{2014}]%
        {campbell2014learning}
\bibfield{author}{\bibinfo{person}{Neill~DF Campbell} {and}
  \bibinfo{person}{Jan Kautz}.} \bibinfo{year}{2014}\natexlab{}.
\newblock \showarticletitle{Learning a manifold of fonts}.
\newblock \bibinfo{journal}{\emph{ACM Transactions on Graphics (TOG)}}
  \bibinfo{volume}{33}, \bibinfo{number}{4} (\bibinfo{year}{2014}),
  \bibinfo{pages}{91}.
\newblock


\bibitem[\protect\citeauthoryear{Chen, Wang, Xu, Jin, and Luo}{Chen
  et~al\mbox{.}}{2019}]%
        {chen2019large}
\bibfield{author}{\bibinfo{person}{Tianlang Chen}, \bibinfo{person}{Zhaowen
  Wang}, \bibinfo{person}{Ning Xu}, \bibinfo{person}{Hailin Jin}, {and}
  \bibinfo{person}{Jiebo Luo}.} \bibinfo{year}{2019}\natexlab{}.
\newblock \showarticletitle{Large-scale Tag-based Font Retrieval with
  Generative Feature Learning}. In \bibinfo{booktitle}{\emph{Proceedings of the
  IEEE International Conference on Computer Vision}}.
  \bibinfo{pages}{9116--9125}.
\newblock


\bibitem[\protect\citeauthoryear{Choi, Matsumura, and Aizawa}{Choi
  et~al\mbox{.}}{2019}]%
        {choi2019assist}
\bibfield{author}{\bibinfo{person}{Saemi Choi}, \bibinfo{person}{Shun
  Matsumura}, {and} \bibinfo{person}{Kiyoharu Aizawa}.}
  \bibinfo{year}{2019}\natexlab{}.
\newblock \showarticletitle{Assist Users' Interactions in Font Search with
  Unexpected but Useful Concepts Generated by Multimodal Learning}. In
  \bibinfo{booktitle}{\emph{Proceedings of the 2019 on International Conference
  on Multimedia Retrieval}}. ACM, \bibinfo{pages}{235--243}.
\newblock


\bibitem[\protect\citeauthoryear{Choi, Choi, Kim, Ha, Kim, and Choo}{Choi
  et~al\mbox{.}}{2018}]%
        {choi2018stargan}
\bibfield{author}{\bibinfo{person}{Yunjey Choi}, \bibinfo{person}{Minje Choi},
  \bibinfo{person}{Munyoung Kim}, \bibinfo{person}{Jung-Woo Ha},
  \bibinfo{person}{Sunghun Kim}, {and} \bibinfo{person}{Jaegul Choo}.}
  \bibinfo{year}{2018}\natexlab{}.
\newblock \showarticletitle{Stargan: Unified generative adversarial networks
  for multi-domain image-to-image translation}. In
  \bibinfo{booktitle}{\emph{Proceedings of the IEEE Conference on Computer
  Vision and Pattern Recognition}}. \bibinfo{pages}{8789--8797}.
\newblock


\bibitem[\protect\citeauthoryear{Gao, Guo, Lian, Tang, and Xiao}{Gao
  et~al\mbox{.}}{2019}]%
        {gao2019artistic}
\bibfield{author}{\bibinfo{person}{Yue Gao}, \bibinfo{person}{Yuan Guo},
  \bibinfo{person}{Zhouhui Lian}, \bibinfo{person}{Yingmin Tang}, {and}
  \bibinfo{person}{Jianguo Xiao}.} \bibinfo{year}{2019}\natexlab{}.
\newblock \showarticletitle{Artistic glyph image synthesis via one-stage
  few-shot learning}.
\newblock \bibinfo{journal}{\emph{ACM Transactions on Graphics (TOG)}}
  \bibinfo{volume}{38}, \bibinfo{number}{6} (\bibinfo{year}{2019}),
  \bibinfo{pages}{185}.
\newblock


\bibitem[\protect\citeauthoryear{Garces, Agarwala, Gutierrez, and
  Hertzmann}{Garces et~al\mbox{.}}{2014}]%
        {garces2014similarity}
\bibfield{author}{\bibinfo{person}{Elena Garces}, \bibinfo{person}{Aseem
  Agarwala}, \bibinfo{person}{Diego Gutierrez}, {and} \bibinfo{person}{Aaron
  Hertzmann}.} \bibinfo{year}{2014}\natexlab{}.
\newblock \showarticletitle{A similarity measure for illustration style}.
\newblock \bibinfo{journal}{\emph{ACM Transactions on Graphics (TOG)}}
  \bibinfo{volume}{33}, \bibinfo{number}{4} (\bibinfo{year}{2014}),
  \bibinfo{pages}{1--9}.
\newblock


\bibitem[\protect\citeauthoryear{Gatys, Ecker, and Bethge}{Gatys
  et~al\mbox{.}}{2016}]%
        {gatys2016image}
\bibfield{author}{\bibinfo{person}{Leon~A Gatys}, \bibinfo{person}{Alexander~S
  Ecker}, {and} \bibinfo{person}{Matthias Bethge}.}
  \bibinfo{year}{2016}\natexlab{}.
\newblock \showarticletitle{Image style transfer using convolutional neural
  networks}. In \bibinfo{booktitle}{\emph{Proceedings of the IEEE conference on
  computer vision and pattern recognition}}. \bibinfo{pages}{2414--2423}.
\newblock


\bibitem[\protect\citeauthoryear{Goodfellow, Pouget-Abadie, Mirza, Xu,
  Warde-Farley, Ozair, Courville, and Bengio}{Goodfellow et~al\mbox{.}}{2014}]%
        {goodfellow2014generative}
\bibfield{author}{\bibinfo{person}{Ian Goodfellow}, \bibinfo{person}{Jean
  Pouget-Abadie}, \bibinfo{person}{Mehdi Mirza}, \bibinfo{person}{Bing Xu},
  \bibinfo{person}{David Warde-Farley}, \bibinfo{person}{Sherjil Ozair},
  \bibinfo{person}{Aaron Courville}, {and} \bibinfo{person}{Yoshua Bengio}.}
  \bibinfo{year}{2014}\natexlab{}.
\newblock \showarticletitle{Generative adversarial nets}. In
  \bibinfo{booktitle}{\emph{Advances in neural information processing
  systems}}. \bibinfo{pages}{2672--2680}.
\newblock


\bibitem[\protect\citeauthoryear{Guo, Lian, Tang, and Xiao}{Guo
  et~al\mbox{.}}{2018}]%
        {guo2018creating}
\bibfield{author}{\bibinfo{person}{Yuan Guo}, \bibinfo{person}{Zhouhui Lian},
  \bibinfo{person}{Yingmin Tang}, {and} \bibinfo{person}{Jianguo Xiao}.}
  \bibinfo{year}{2018}\natexlab{}.
\newblock \showarticletitle{Creating New Chinese Fonts based on Manifold
  Learning and Adversarial Networks.}. In
  \bibinfo{booktitle}{\emph{Eurographics (Short Papers)}}.
  \bibinfo{pages}{61--64}.
\newblock


\bibitem[\protect\citeauthoryear{He, Zhang, Ren, and Sun}{He
  et~al\mbox{.}}{2016}]%
        {he2016deep}
\bibfield{author}{\bibinfo{person}{Kaiming He}, \bibinfo{person}{Xiangyu
  Zhang}, \bibinfo{person}{Shaoqing Ren}, {and} \bibinfo{person}{Jian Sun}.}
  \bibinfo{year}{2016}\natexlab{}.
\newblock \showarticletitle{Deep residual learning for image recognition}. In
  \bibinfo{booktitle}{\emph{Proceedings of the IEEE conference on computer
  vision and pattern recognition}}. \bibinfo{pages}{770--778}.
\newblock


\bibitem[\protect\citeauthoryear{He, Zuo, Kan, Shan, and Chen}{He
  et~al\mbox{.}}{2019}]%
        {he2019attgan}
\bibfield{author}{\bibinfo{person}{Zhenliang He}, \bibinfo{person}{Wangmeng
  Zuo}, \bibinfo{person}{Meina Kan}, \bibinfo{person}{Shiguang Shan}, {and}
  \bibinfo{person}{Xilin Chen}.} \bibinfo{year}{2019}\natexlab{}.
\newblock \showarticletitle{Attgan: Facial attribute editing by only changing
  what you want}.
\newblock \bibinfo{journal}{\emph{IEEE Transactions on Image Processing}}
  (\bibinfo{year}{2019}).
\newblock


\bibitem[\protect\citeauthoryear{Heusel, Ramsauer, Unterthiner, Nessler, and
  Hochreiter}{Heusel et~al\mbox{.}}{2017}]%
        {heusel2017gans}
\bibfield{author}{\bibinfo{person}{Martin Heusel}, \bibinfo{person}{Hubert
  Ramsauer}, \bibinfo{person}{Thomas Unterthiner}, \bibinfo{person}{Bernhard
  Nessler}, {and} \bibinfo{person}{Sepp Hochreiter}.}
  \bibinfo{year}{2017}\natexlab{}.
\newblock \showarticletitle{Gans trained by a two time-scale update rule
  converge to a local nash equilibrium}. In \bibinfo{booktitle}{\emph{Advances
  in Neural Information Processing Systems}}. \bibinfo{pages}{6626--6637}.
\newblock


\bibitem[\protect\citeauthoryear{Isola, Zhu, Zhou, and Efros}{Isola
  et~al\mbox{.}}{2017}]%
        {isola2017image}
\bibfield{author}{\bibinfo{person}{Phillip Isola}, \bibinfo{person}{Jun-Yan
  Zhu}, \bibinfo{person}{Tinghui Zhou}, {and} \bibinfo{person}{Alexei~A
  Efros}.} \bibinfo{year}{2017}\natexlab{}.
\newblock \showarticletitle{Image-to-image translation with conditional
  adversarial networks}. In \bibinfo{booktitle}{\emph{Proceedings of the IEEE
  conference on computer vision and pattern recognition}}.
  \bibinfo{pages}{1125--1134}.
\newblock


\bibitem[\protect\citeauthoryear{Jiang, Lian, Tang, and Xiao}{Jiang
  et~al\mbox{.}}{2017}]%
        {jiang2017dcfont}
\bibfield{author}{\bibinfo{person}{Yue Jiang}, \bibinfo{person}{Zhouhui Lian},
  \bibinfo{person}{Yingmin Tang}, {and} \bibinfo{person}{Jianguo Xiao}.}
  \bibinfo{year}{2017}\natexlab{}.
\newblock \showarticletitle{DCFont: an end-to-end deep Chinese font generation
  system}. In \bibinfo{booktitle}{\emph{SIGGRAPH Asia 2017 Technical Briefs}}.
  ACM, \bibinfo{pages}{22}.
\newblock


\bibitem[\protect\citeauthoryear{Jiang, Lian, Tang, and Xiao}{Jiang
  et~al\mbox{.}}{2019}]%
        {jiang2019scfont}
\bibfield{author}{\bibinfo{person}{Yue Jiang}, \bibinfo{person}{Zhouhui Lian},
  \bibinfo{person}{Yingmin Tang}, {and} \bibinfo{person}{Jianguo Xiao}.}
  \bibinfo{year}{2019}\natexlab{}.
\newblock \showarticletitle{SCFont: Structure-guided Chinese Font Generation
  via Deep Stacked Networks}.
\newblock  (\bibinfo{year}{2019}).
\newblock


\bibitem[\protect\citeauthoryear{{Kingma} and {Ba}}{{Kingma} and {Ba}}{2015}]%
        {kingma2015adam}
\bibfield{author}{\bibinfo{person}{Diederik~P. {Kingma}} {and}
  \bibinfo{person}{Jimmy~Lei {Ba}}.} \bibinfo{year}{2015}\natexlab{}.
\newblock \showarticletitle{Adam: A Method for Stochastic Optimization}.
\newblock \bibinfo{journal}{\emph{ICLR}} (\bibinfo{year}{2015}).
\newblock


\bibitem[\protect\citeauthoryear{Lian, Zhao, Chen, and Xiao}{Lian
  et~al\mbox{.}}{2018}]%
        {lian2018easyfont}
\bibfield{author}{\bibinfo{person}{Zhouhui Lian}, \bibinfo{person}{Bo Zhao},
  \bibinfo{person}{Xudong Chen}, {and} \bibinfo{person}{Jianguo Xiao}.}
  \bibinfo{year}{2018}\natexlab{}.
\newblock \showarticletitle{EasyFont: A Style Learning-Based System to Easily
  Build Your Large-Scale Handwriting Fonts}.
\newblock \bibinfo{journal}{\emph{ACM Transactions on Graphics (TOG)}}
  \bibinfo{volume}{38}, \bibinfo{number}{1} (\bibinfo{year}{2018}),
  \bibinfo{pages}{6}.
\newblock


\bibitem[\protect\citeauthoryear{Liu, Ding, Xia, Liu, Ding, Zuo, and Wen}{Liu
  et~al\mbox{.}}{2019}]%
        {liu2019stgan}
\bibfield{author}{\bibinfo{person}{Ming Liu}, \bibinfo{person}{Yukang Ding},
  \bibinfo{person}{Min Xia}, \bibinfo{person}{Xiao Liu}, \bibinfo{person}{Errui
  Ding}, \bibinfo{person}{Wangmeng Zuo}, {and} \bibinfo{person}{Shilei Wen}.}
  \bibinfo{year}{2019}\natexlab{}.
\newblock \showarticletitle{STGAN: A Unified Selective Transfer Network for
  Arbitrary Image Attribute Editing}. In \bibinfo{booktitle}{\emph{Proceedings
  of the IEEE Conference on Computer Vision and Pattern Recognition}}.
  \bibinfo{pages}{3673--3682}.
\newblock


\bibitem[\protect\citeauthoryear{Liu, Luo, Wang, and Tang}{Liu
  et~al\mbox{.}}{2015}]%
        {liu2015faceattributes}
\bibfield{author}{\bibinfo{person}{Ziwei Liu}, \bibinfo{person}{Ping Luo},
  \bibinfo{person}{Xiaogang Wang}, {and} \bibinfo{person}{Xiaoou Tang}.}
  \bibinfo{year}{2015}\natexlab{}.
\newblock \showarticletitle{Deep Learning Face Attributes in the Wild}. In
  \bibinfo{booktitle}{\emph{Proceedings of International Conference on Computer
  Vision (ICCV)}}.
\newblock


\bibitem[\protect\citeauthoryear{Lopes, Ha, Eck, and Shlens}{Lopes
  et~al\mbox{.}}{2019}]%
        {lopes2019learned}
\bibfield{author}{\bibinfo{person}{Raphael~Gontijo Lopes},
  \bibinfo{person}{David Ha}, \bibinfo{person}{Douglas Eck}, {and}
  \bibinfo{person}{Jonathon Shlens}.} \bibinfo{year}{2019}\natexlab{}.
\newblock \showarticletitle{A Learned Representation for Scalable Vector
  Graphics}.
\newblock \bibinfo{journal}{\emph{arXiv preprint arXiv:1904.02632}}
  (\bibinfo{year}{2019}).
\newblock


\bibitem[\protect\citeauthoryear{Lun, Kalogerakis, and Sheffer}{Lun
  et~al\mbox{.}}{2015}]%
        {lun2015elements}
\bibfield{author}{\bibinfo{person}{Zhaoliang Lun}, \bibinfo{person}{Evangelos
  Kalogerakis}, {and} \bibinfo{person}{Alla Sheffer}.}
  \bibinfo{year}{2015}\natexlab{}.
\newblock \showarticletitle{Elements of style: learning perceptual shape style
  similarity}.
\newblock \bibinfo{journal}{\emph{ACM Transactions on graphics (TOG)}}
  \bibinfo{volume}{34}, \bibinfo{number}{4} (\bibinfo{year}{2015}),
  \bibinfo{pages}{1--14}.
\newblock


\bibitem[\protect\citeauthoryear{Lyu, Bai, Yao, Zhu, Huang, and Liu}{Lyu
  et~al\mbox{.}}{2017}]%
        {lyu2017auto}
\bibfield{author}{\bibinfo{person}{Pengyuan Lyu}, \bibinfo{person}{Xiang Bai},
  \bibinfo{person}{Cong Yao}, \bibinfo{person}{Zhen Zhu},
  \bibinfo{person}{Tengteng Huang}, {and} \bibinfo{person}{Wenyu Liu}.}
  \bibinfo{year}{2017}\natexlab{}.
\newblock \showarticletitle{Auto-encoder guided GAN for Chinese calligraphy
  synthesis}. In \bibinfo{booktitle}{\emph{2017 14th IAPR International
  Conference on Document Analysis and Recognition (ICDAR)}},
  Vol.~\bibinfo{volume}{1}. IEEE, \bibinfo{pages}{1095--1100}.
\newblock


\bibitem[\protect\citeauthoryear{Ma, Huang, Sheffer, Kalogerakis, and Wang}{Ma
  et~al\mbox{.}}{2014}]%
        {ma2014analogy}
\bibfield{author}{\bibinfo{person}{Chongyang Ma}, \bibinfo{person}{Haibin
  Huang}, \bibinfo{person}{Alla Sheffer}, \bibinfo{person}{Evangelos
  Kalogerakis}, {and} \bibinfo{person}{Rui Wang}.}
  \bibinfo{year}{2014}\natexlab{}.
\newblock \showarticletitle{Analogy-driven 3D style transfer}. In
  \bibinfo{booktitle}{\emph{Computer Graphics Forum}},
  Vol.~\bibinfo{volume}{33}. Wiley Online Library, \bibinfo{pages}{175--184}.
\newblock


\bibitem[\protect\citeauthoryear{Mechrez, Talmi, and Zelnik-Manor}{Mechrez
  et~al\mbox{.}}{2018}]%
        {mechrez2018contextual}
\bibfield{author}{\bibinfo{person}{Roey Mechrez}, \bibinfo{person}{Itamar
  Talmi}, {and} \bibinfo{person}{Lihi Zelnik-Manor}.}
  \bibinfo{year}{2018}\natexlab{}.
\newblock \showarticletitle{The contextual loss for image transformation with
  non-aligned data}. In \bibinfo{booktitle}{\emph{Proceedings of the European
  Conference on Computer Vision (ECCV)}}. \bibinfo{pages}{768--783}.
\newblock


\bibitem[\protect\citeauthoryear{O'Donovan, L{\=\i}beks, Agarwala, and
  Hertzmann}{O'Donovan et~al\mbox{.}}{2014}]%
        {o2014exploratory}
\bibfield{author}{\bibinfo{person}{Peter O'Donovan}, \bibinfo{person}{J{\=a}nis
  L{\=\i}beks}, \bibinfo{person}{Aseem Agarwala}, {and} \bibinfo{person}{Aaron
  Hertzmann}.} \bibinfo{year}{2014}\natexlab{}.
\newblock \showarticletitle{Exploratory font selection using crowdsourced
  attributes}.
\newblock \bibinfo{journal}{\emph{ACM Transactions on Graphics (TOG)}}
  \bibinfo{volume}{33}, \bibinfo{number}{4} (\bibinfo{year}{2014}),
  \bibinfo{pages}{92}.
\newblock


\bibitem[\protect\citeauthoryear{Salimans, Goodfellow, Zaremba, Cheung,
  Radford, and Chen}{Salimans et~al\mbox{.}}{2016}]%
        {salimans2016improved}
\bibfield{author}{\bibinfo{person}{Tim Salimans}, \bibinfo{person}{Ian
  Goodfellow}, \bibinfo{person}{Wojciech Zaremba}, \bibinfo{person}{Vicki
  Cheung}, \bibinfo{person}{Alec Radford}, {and} \bibinfo{person}{Xi Chen}.}
  \bibinfo{year}{2016}\natexlab{}.
\newblock \showarticletitle{Improved techniques for training gans}. In
  \bibinfo{booktitle}{\emph{Advances in neural information processing
  systems}}. \bibinfo{pages}{2234--2242}.
\newblock


\bibitem[\protect\citeauthoryear{Serrano, Gutierrez, Myszkowski, Seidel, and
  Masia}{Serrano et~al\mbox{.}}{2018}]%
        {serrano2018intuitive}
\bibfield{author}{\bibinfo{person}{Ana Serrano}, \bibinfo{person}{Diego
  Gutierrez}, \bibinfo{person}{Karol Myszkowski}, \bibinfo{person}{Hans-Peter
  Seidel}, {and} \bibinfo{person}{Belen Masia}.}
  \bibinfo{year}{2018}\natexlab{}.
\newblock \showarticletitle{An intuitive control space for material
  appearance}.
\newblock \bibinfo{journal}{\emph{arXiv preprint arXiv:1806.04950}}
  (\bibinfo{year}{2018}).
\newblock


\bibitem[\protect\citeauthoryear{Ulyanov, Vedaldi, and Lempitsky}{Ulyanov
  et~al\mbox{.}}{2016}]%
        {ulyanov2016instance}
\bibfield{author}{\bibinfo{person}{Dmitry Ulyanov}, \bibinfo{person}{Andrea
  Vedaldi}, {and} \bibinfo{person}{Victor Lempitsky}.}
  \bibinfo{year}{2016}\natexlab{}.
\newblock \showarticletitle{Instance normalization: The missing ingredient for
  fast stylization}.
\newblock \bibinfo{journal}{\emph{arXiv preprint arXiv:1607.08022}}
  (\bibinfo{year}{2016}).
\newblock


\bibitem[\protect\citeauthoryear{Wang, Yang, Jin, Brandt, Shechtman, Agarwala,
  Wang, Song, Hsieh, Kong, et~al\mbox{.}}{Wang et~al\mbox{.}}{2015}]%
        {wang2015deepfont}
\bibfield{author}{\bibinfo{person}{Zhangyang Wang}, \bibinfo{person}{Jianchao
  Yang}, \bibinfo{person}{Hailin Jin}, \bibinfo{person}{Jonathan Brandt},
  \bibinfo{person}{Eli Shechtman}, \bibinfo{person}{Aseem Agarwala},
  \bibinfo{person}{Zhaowen Wang}, \bibinfo{person}{Yuyan Song},
  \bibinfo{person}{Joseph Hsieh}, \bibinfo{person}{Sarah Kong},
  {et~al\mbox{.}}} \bibinfo{year}{2015}\natexlab{}.
\newblock \showarticletitle{Deepfont: A system for font recognition and
  similarity}. In \bibinfo{booktitle}{\emph{23rd ACM International Conference
  on Multimedia, MM 2015}}. Association for Computing Machinery, Inc,
  \bibinfo{pages}{813--814}.
\newblock


\bibitem[\protect\citeauthoryear{Woo, Park, Lee, and So~Kweon}{Woo
  et~al\mbox{.}}{2018}]%
        {woo2018cbam}
\bibfield{author}{\bibinfo{person}{Sanghyun Woo}, \bibinfo{person}{Jongchan
  Park}, \bibinfo{person}{Joon-Young Lee}, {and} \bibinfo{person}{In
  So~Kweon}.} \bibinfo{year}{2018}\natexlab{}.
\newblock \showarticletitle{Cbam: Convolutional block attention module}. In
  \bibinfo{booktitle}{\emph{Proceedings of the European Conference on Computer
  Vision (ECCV)}}. \bibinfo{pages}{3--19}.
\newblock


\bibitem[\protect\citeauthoryear{Wu, Lin, Chang, Chang, and Liao}{Wu
  et~al\mbox{.}}{2019}]%
        {wu2019relgan}
\bibfield{author}{\bibinfo{person}{Po-Wei Wu}, \bibinfo{person}{Yu-Jing Lin},
  \bibinfo{person}{Che-Han Chang}, \bibinfo{person}{Edward~Y Chang}, {and}
  \bibinfo{person}{Shih-Wei Liao}.} \bibinfo{year}{2019}\natexlab{}.
\newblock \showarticletitle{Relgan: Multi-domain image-to-image translation via
  relative attributes}. In \bibinfo{booktitle}{\emph{Proceedings of the IEEE
  International Conference on Computer Vision}}. \bibinfo{pages}{5914--5922}.
\newblock


\bibitem[\protect\citeauthoryear{Xu, Zhang, Huang, Zhang, Gan, Huang, and
  He}{Xu et~al\mbox{.}}{2018}]%
        {xu2018attngan}
\bibfield{author}{\bibinfo{person}{Tao Xu}, \bibinfo{person}{Pengchuan Zhang},
  \bibinfo{person}{Qiuyuan Huang}, \bibinfo{person}{Han Zhang},
  \bibinfo{person}{Zhe Gan}, \bibinfo{person}{Xiaolei Huang}, {and}
  \bibinfo{person}{Xiaodong He}.} \bibinfo{year}{2018}\natexlab{}.
\newblock \showarticletitle{Attngan: Fine-grained text to image generation with
  attentional generative adversarial networks}. In
  \bibinfo{booktitle}{\emph{Proceedings of the IEEE Conference on Computer
  Vision and Pattern Recognition}}. \bibinfo{pages}{1316--1324}.
\newblock


\bibitem[\protect\citeauthoryear{Zhang, Isola, Efros, Shechtman, and
  Wang}{Zhang et~al\mbox{.}}{2018a}]%
        {zhang2018unreasonable}
\bibfield{author}{\bibinfo{person}{Richard Zhang}, \bibinfo{person}{Phillip
  Isola}, \bibinfo{person}{Alexei~A Efros}, \bibinfo{person}{Eli Shechtman},
  {and} \bibinfo{person}{Oliver Wang}.} \bibinfo{year}{2018}\natexlab{a}.
\newblock \showarticletitle{The unreasonable effectiveness of deep features as
  a perceptual metric}. In \bibinfo{booktitle}{\emph{Proceedings of the IEEE
  Conference on Computer Vision and Pattern Recognition}}.
  \bibinfo{pages}{586--595}.
\newblock


\bibitem[\protect\citeauthoryear{Zhang, Li, Li, Wang, Zhong, and Fu}{Zhang
  et~al\mbox{.}}{2018b}]%
        {zhang2018image}
\bibfield{author}{\bibinfo{person}{Yulun Zhang}, \bibinfo{person}{Kunpeng Li},
  \bibinfo{person}{Kai Li}, \bibinfo{person}{Lichen Wang},
  \bibinfo{person}{Bineng Zhong}, {and} \bibinfo{person}{Yun Fu}.}
  \bibinfo{year}{2018}\natexlab{b}.
\newblock \showarticletitle{Image super-resolution using very deep residual
  channel attention networks}. In \bibinfo{booktitle}{\emph{Proceedings of the
  European Conference on Computer Vision (ECCV)}}. \bibinfo{pages}{286--301}.
\newblock


\end{thebibliography}

\end{document}